\documentclass[lettersize,journal]{IEEEtran}
\usepackage{amsmath,amsfonts}
\usepackage{algorithmic}
\usepackage{algorithm}
\usepackage{array}
\usepackage{textcomp}
\usepackage{stfloats}
\usepackage{url}
\usepackage{verbatim}
\usepackage{graphicx}
\usepackage{cite}
\usepackage{bm}
\usepackage{amssymb}
\usepackage{threeparttable}
\usepackage{float}
\usepackage{booktabs}
\usepackage{subfigure}
\usepackage{multirow}
\usepackage{diagbox}
\usepackage{subfigure}

\begin{document}

\title{Augmenting Prototype Network with TransMix for Few-shot Hyperspectral Image Classification}

\author{Chun Liu, Longwei Yang, Dongmei Dong, Zheng Li, Wei Yang, Zhigang Han, and Jiayao Wang 
\thanks{This paper was produced by the IEEE Publication Technology Group. They are in Piscataway, NJ.}
\thanks{Manuscript received xxx, 2023; revised xxx xx, 2023.}
\thanks{Chun Liu, Longwei Yang, Zheng Li and Wei Yang are with School of Computer and Information Engineering, Henan Key Laboratory of Big Data Analysis and Processing, Henan Engineering Laboratory of Spatial Information Processing and Henan Industrial Technology Academy of Spatio-Temporal Big Data, Henan University, Zhengzhou 450046, China (e-mail: liuchun@henu.edu.cn; 104754211314@henu.edu.cn; lizheng@henu.edu.cn; yangwei@henu.edu.cn).}
\thanks{Dongmei Dong is with Beijing Electro-Mechanical Engineering Institute, Beijing 100074, China (e-mail: dongdongmeiangel@163.com);}
\thanks{Zhigang Han and Jiayao Wang are with College of Geography and Environmental Science and Henan Industrial Technology Academy of Spatio-Temporal Big Data, Henan University, Zhengzhou 450046, China (e-mail: zghan@henu.edu.cn, wjy@henu.edu.cn);}
}

\markboth{Journal of \LaTeX\ Class Files,~Vol.~xx, No.~xx, xxx ~202x}%
{Shell \MakeLowercase{\textit{et al.}}: A Sample Article Using IEEEtran.cls for IEEE Journals}

\maketitle

\begin{abstract}
	
Few-shot hyperspectral image classification aims to identify the classes of each pixel in the images by only marking few of these pixels. And in order to obtain the spatial-spectral joint features of each pixel, the fixed-size patches centering around each pixel are often used for classification. However, observing the classification results of existing methods, we found that boundary patches corresponding to the pixels which are located at the boundary of the objects in the hyperspectral images, are hard to classify. These boundary patchs are mixed with multi-class spectral information. Inspired by this, we propose to augment the prototype network with TransMix for few-shot hyperspectrial image classification(APNT). While taking the prototype network as the backbone, it adopts the transformer as feature extractor to learn the pixel-to-pixel relation and pay different attentions to different pixels. At the same time, instead of directly using the patches which are cut from the hyperspectral images for training, it randomly mixs up two patches to imitate the boundary patches and uses the synthetic patches to train the model, with the aim to enlarge the number of hard training samples and enhance their diversity.  And by following the data agumentation technique TransMix, the attention returned by the transformer is also used to mix up the labels of two patches to generate better labels for synthetic patches. Compared with existing methods, the proposed method has demonstrated sate of the art performance and better robustness for few-shot hyperspectral image classification in our experiments. All the codes are available at https://github.com/HENULWY/APNT.
	
\end{abstract}


\section{Introduction}
\IEEEPARstart{H}{yperspectral} images (HSIs) capture radiation information of ground objects over dozens or even hundreds of continuous spectral channels. It can obtain spectral features that reflect the unique characteristics of the targets. Compared to natural images with only RGB three channels, HSIs integrate spatial and spectral features of the ground objects, which can capture more subtle differences between them. Due to this advantage, HSIs have great application value in the fields such as  environmental monitoring\cite{1} and resource utilization and management\cite{3,4}.

HSI classification, which is one fundamental task for HSI applications,  is to classify each pixel of HSIs and predict their classes. Because the spectral vectors behind each pixel of HSIs contain rich radiation information, the works for HSI classification in early stage focus mainly on the spectral features, and directly take these spectral vectors as the samples to be classified \cite{5,6}. But caused by the factors such as illumination and atmosphere, the phenomenon of same objects with different spectrum and different objects with same spectrum is widely present in HSIs. Therefore, later methods for HSIs classification pay much attention to integrating the spatial features with the spectral features of each pixel \cite{7,8}, where the spatial features provide additional useful information about the shape, context, and layout around each pixel. To capture the spatial features,  the fixed-size (e.g., $9\times 9$) patches  centering around each pixel are often generated from HSIs and then taken as the samples to be classified.

In recent years, with the powerful feature extraction ability and a great success in a series of fields, deep learning  has been widely used for HSI classification \cite{9,10,11}. Based on these typical deep learning models of convolutional neural networks (CNN), recurrent neural networks (RNN) and graph neural networks (GNN), many deep learning methods have been designed to obtain more  distinguishable spatial and spectral features from HSIs.  Some methods adopt the two branches architecture, which uses different neural networks to extract the spatial and spectral features respectively and then fuses them together \cite{12,13,14}. At the same time,  many methods strive to use a single feature extractor such as 3DCNN to capture the spatial-spectrial joint features \cite{15,16}. 

In light of that deep learning  methods often  require a large number of labeled training samples and it is expensive and time-consuming to manually mark these samples, few-shot learning methods have also aroused lots of interest recently \cite{18, 19,20,21,22,23}. Few-shot learning,  which is a branch of deep learning,  aims to learn to identify some classes of samples by  marking only few of them, e.g.,  three or five samples per class. This kind of ability has been viewed as one skill possessed by humans. With the aids of available abundant   labeld samples, current few-shot learning methods mainly follow the way of transfer learning, that is,  acquire prior knowledge from  available abundant labeled samples and then transfer the knowledge to target tasks which contain only few labeled samples. To transfer knowledge effectively,  few-shot learning methods often adopt the meta-learning technique which constructs lots of similar tasks by imitating the target tasks and uses the constructed tasks to train a model that can easily adapt to target tasks. Based on typical few-shot learning models such as prototype network \cite{18},  many few-shot  methods have been designed for HSI classification. For example, DFSL \cite{24} took the 3D residual network as feature extractor to extract better spatial-spectral joint features from HSIs. Different from DFSL, SSPN \cite{25} used the local pattern coding technique to  combine the spatial and spectral information, and then applied  1D convolutional neural network as extractor to obtain features. Instead of using the Euclidean distance,  HSEMD-Net \cite{26} used the Earth Mover distance  to learn prototype representations for each hyperspectral class. Meanwhile, RL-Net \cite{27} followed the relational network \cite{28} which is an improved version of prototype network for few-shot HSI classification.

When classifying HSIs under few-shot setting, early methods usually assume that the samples from which the  prior knowledge are learned and the samples of the target tasks  are coming from the same domain. This often means that these samples should be captured by the same sensors and in the same environment.  To relax this assumption, a set of crosss-domain few-shot methods have been further proposed for HSI classification.  Their purposes are to enhance model's cross-domain generalization ability. For example,  DCFSL \cite{29}  augmented DFSL method with a domain discriminator to obtain domain-independent features;  SSFT \cite{30} changed the feature distribution by a feature-wise transformation module with the aim to obtain generalized features; CMFSL \cite{31} used the  task-adapted class-covariance metric to obtain better features under few-shot setting; Gia-CFSL \cite{32} enhanced prototype network with domain alignment strategy to increase domain adaptation. By using supervised contrastive learning,  RPCL\cite{33} imposed triple constraints on prototypes of the support set to stabilize and refine the prototypes. Moreover, MRLF proposed to learn task-specific relations between samples, including the contrastive and affinitive relations, to futher improve the feature discriminability \cite{34}. In addition, considering the availability of the large number of natural images, HFSL \cite{35} took natural image datasets for pre-training to obtain prior classification knowledge, allowing the model to better distinguish hyperspectral samples.

\begin{figure}[htbp]
	\centering
	\subfigure[]{
		\includegraphics[width=2.5cm,height=3.5cm]{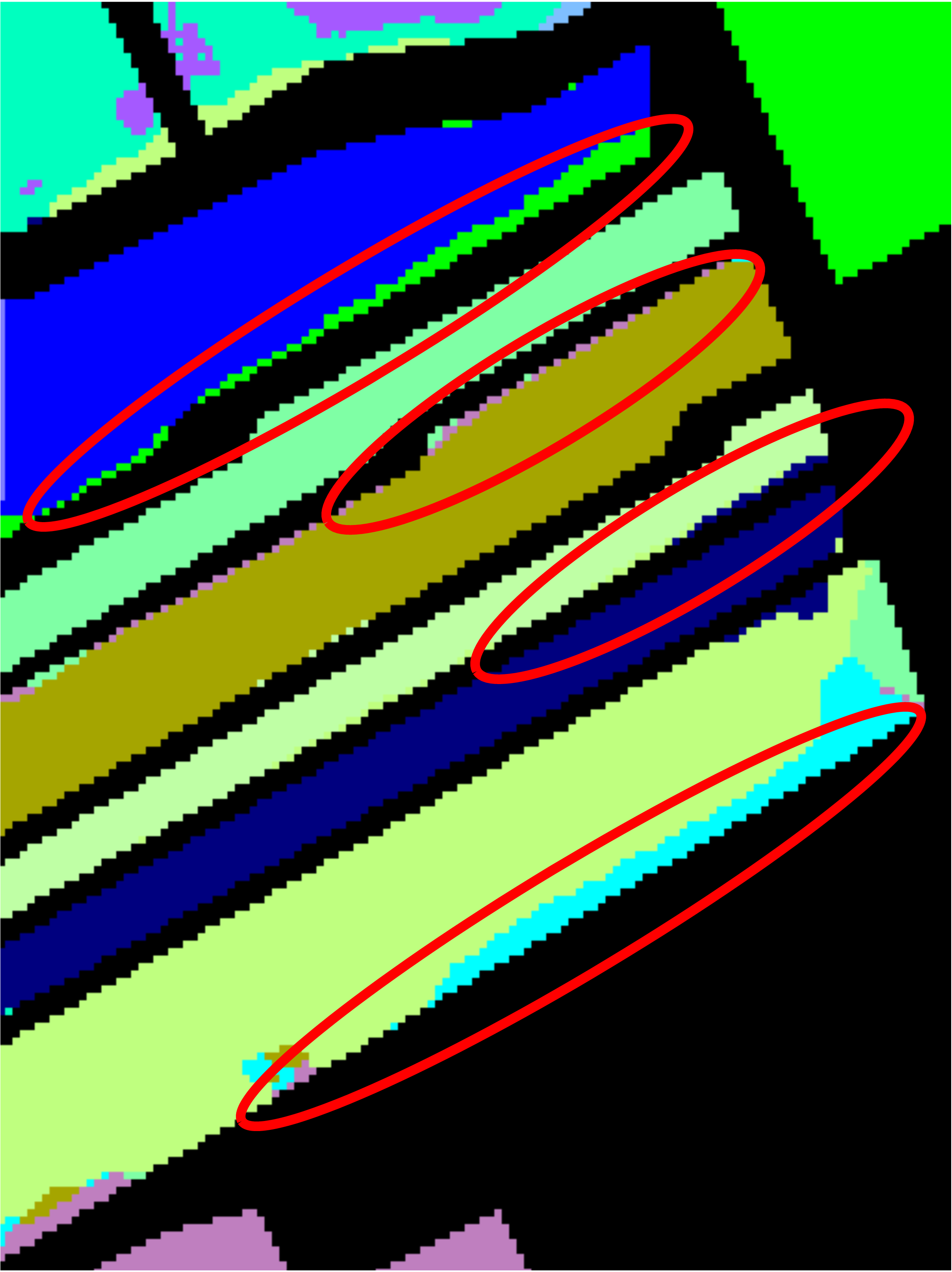}
	}
	\subfigure[]{
		\includegraphics[width=3.8cm,height=3.6cm]{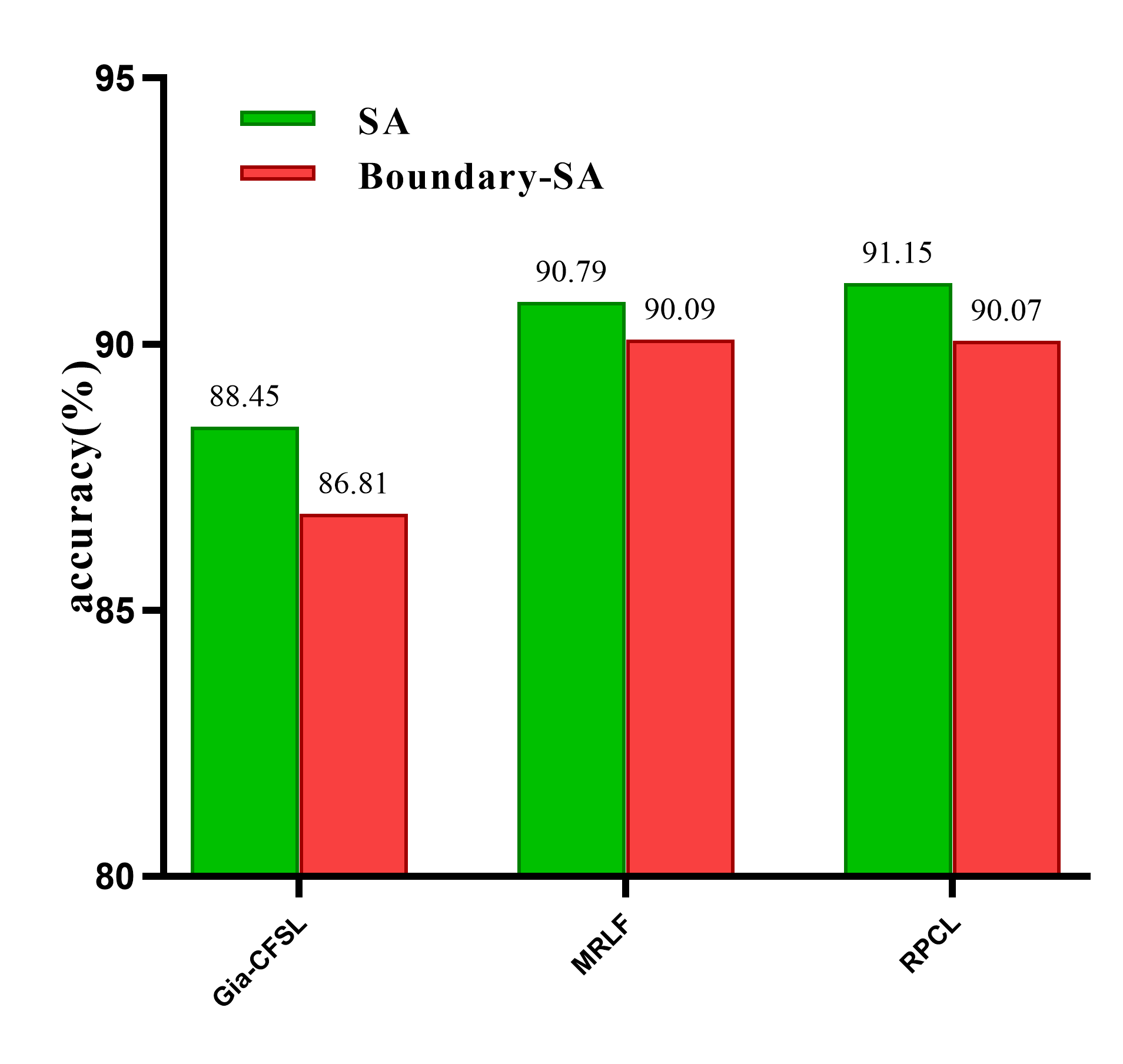}
	}
	\caption{(a)The classfication map indicating incorrect classification often occurs at boundary pixels; and (b) The overal accuracy of current methods on SA dataset (green color) and on these boundary pixels in SA dataset (red color).}
	\label{Boundary-Pixel}
\end{figure}

However, while taking the fixed-size patches around each pixel as samples to obtain spatial-spectral joint features,  current few-shot research works for HSI classification, including these cross-domain methods,  have paid less attention to the difference between the patches around the boundary pixels and the others.  As shown in Fig. \ref{Boundary-Pixel}, each ground object in HSIs occupies a part of the images, which covers a set of pixels. The boundary patches built for the boundary pixels of the object will be different from these patches for the pixels located far from the boundary. These boundary patches will constitute the pixels belonging to the adjacent ground objects, which will be also mixed with multi-class spectral information. This means that the classification results of these boundary patches will be affected by these pixels from adjacent objects and the different pixels in the patches will have different importance. Generally, these boundary patches are the hard samples to be classified. Our primary experimental results shown in Fig. \ref{Boundary-Pixel} have also indcated that current methods have shown lower performance for these boundary patches. What's worse, these boundary patches only constitute a minority in the training samples, so they have less contributions to improve the robustness of the classification model.

To address above issue,  we augment the prototype network with TransMix for few-shot HSI classification (APNT) in this paper.  In light of that boundary patches are the samples mixed by the pixels from different objects, our main idea is to randomly mix up two different training samples online and use the synthetic samples to train the model. The typical few-shot learning model prototype network, which takes the few labeled samples as supports to caculate the class prototypes and predicts the unlabeled query samples according to their distance to the prototypes, is adopted as the backbone of our proposed network. To learn about the different importantce of the pixels in the patches,  the proposed network enhance the prototype network by using transformer as the feature extractor. Through treating each patch as a sequence of pixels which are the spectral vectors, this can enable the model to pay different attention to different pixels in the patche. Moreover, inspired by TransMix \cite{36} which is a transformer based mix-up technique, the attention returned from transformer will be further used to derive the labels for the synthetic samples, to  better reflect the contributions of each sample to the synthetic samples. In sumary, the proposed method is expected to enhance the robustness of prototype network when classifying these boundary pixels in HSIs. The main contributions of this work are as follows.

\begin{enumerate}
	
	\item{We propose to randomly mix up two HSI patches by imitating the boundary patches which are centering around the boundary pixels of objects in HSIs and use these synthetic patches to train the model, in order to enlarge the number of hard training samples and enhance their diversity for few-shot HSI classification.}
	
	\item{A lightweight TransMix based prototype network is designed for mixing up patches online for few-shot HSI classification, which adopts transformer as feature extractor to pay different attention to different pixels in the patches, and takes the attention to mix up the labels of two patches to generate better labels for synthetic patches.}

	\item {Extensive experiments have been done and the proposed method has shown the state of art performance when compared with latest methods. The experiments also reveal that without the pre-training on existing datasets, the proposed method also has comparable performance by using only these few labeled samples in the target tasks. This will make it more convenient for its use in a range of applications.}
\end{enumerate}

The remainder of this paper is structured as follows. Section 2 details the proposed method. Section 3 describes the datasets, the design and the results of the experiments. Finally, the conclusions are in section 4.

\section{The proposed method}

In this section, we state the problem of few-shot HSI classification, and detail the proposed method. 

\begin{figure*}[ht]
	\centering
	\includegraphics[width=16cm,height=5.4cm]{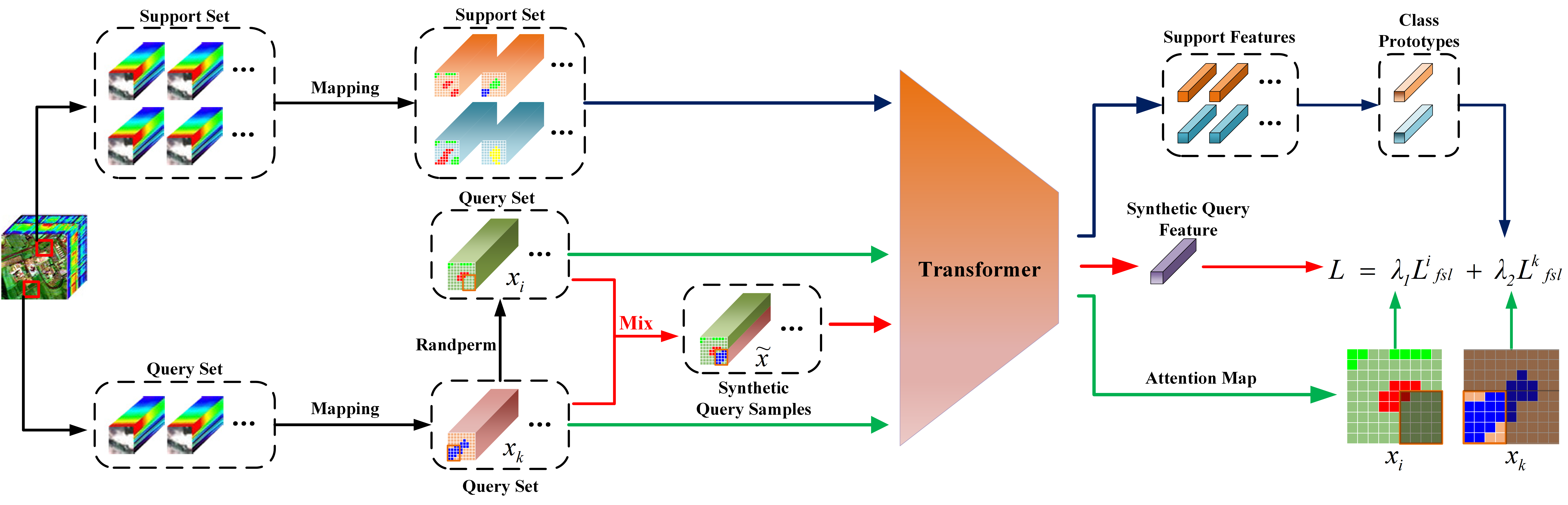}
	\caption{The workflow of APNT model. The patches sampled from the datasets are divided into Support set and Query set. Before passing through transformer to extract the spatial-spectral joint features, the synthetic query samples are generated by randomly mixing up two query samples in the query set. Once the features are obtained from transformer, the class prototypes will be derived as the mean of these support features from the same classes. In the meanwhile, the labels of these synthetic query samples are generated by mixing up the labels of two query samples according to the attention returned by transformer. Finally, the cross entropy loss is caculated with the synthetic labels.}
	\label{workflow}
\end{figure*}

\subsection{Problem Statement}
For the problem of few-shot HSI classification, there are two datasets given: source dataset and target dataset. In practical application, all the samples in the source dataset are labeled, which are used for training the classification model. While, there are only few labeled samples in the target dataset. The samples are the pixels of the HSIs. In order to fully preserve spatial information, the fixed-size patches centering around each pixel instead of the pixeles themselves are often treated as the samples. The few-shot learning task is to predict the classes of these unlabeled samples in the target dataset by using all these labeled samples in source dataset and target dataset. Nevertheless,  another dataset in which all the samples are also labeled is often selected as the target dataset in the research environment, to facilitate the performance evaluation of the few-shot learning methods. Such dataset can be also called testing dataset. 

To elaborate the problem formally, it is often assumed that there are $C_s$ classses in the source dataset $D_s$ and $C_t$ classses in the target dataset $D_t$, where $C_t$ is smaller than $C_s$. To imitate the setting of target dataset,  lots of the tasks will be constructed by randomly selecting $C \times (K + M)$ samples each time, and used to train the classification model in an episode manner. The $C$ refers to the number of classes selected, which is often set to the number of classes in the target dataset. $K$ is the number of few labeled samples per class, and $M$ is the number of query samples per class to be classified, where $K$ is smaller than $M$. The set consisiting of the $C\times K$ labeled samples is often called support set $ \{ ({x_i},{y_i})\} _{i = 1}^{{C \times K}}$, while these $C \times M $ query samples will form the query set $ \{ ({x_i},{y_i})\} _{i = 1}^{{C \times M }}$. Such few-shot tasks are often called the $C$-way $K$-shot tasks, whose purpose are to predict the classes of these query samples by using these support samples.

\subsection{The Model Architecture and Learning Process}

The architecture of the proposed APNT method and the flowchart of the learning process is shown in Fig. \ref{workflow}. APNT model mainly consists of three components: query sample mixing, transformer based feature extractor, and TransMix based few-shot learning loss. Given an input task consisting of the support and query set, the query sample mixing module will mix up each query sample with another randomly selected query sample in the way of CutMix\cite{37}, and output the synthetic query samples. Subsequently, the transformer based feature extractor extracts the embedding features of the support samples and the synthetic query samples, and also returns the attention maps reflecting the importance of each pixel of the original query samples. Following that, the mean of the embedding features of the support samples in the same classes will be caculated as the prototpyes of each class. In the meanwhile, the TransMix based few-shot learning loss module computes the cross entropy loss of the synthetic query samples according to their synthetic labels which are derived by using the attention maps returned by transformer.

Similar with the work of RPCL\cite{33}, We adopt the fine-tuning strategy to train the network. That is, we first pre-train the network by using the tasks constructed from $D_s$, and then fine-tune it by using the tasks from $D_t$. This also means there are some labeled samples selected from $D_t$ to participate in the training. Consistent with previous works, there are five samples per class selected, which are augmented to 200 samples per class in our experiments. The tasks from $D_t$ will be constructed from these augmented smaples. Finally, the remaining samples in the target dataset $D_t$ are used for testing. And after extracting sample features with the trained transformer, the classes of these query samples are pedicted by using KNN algorithm during testing. It is worth noting that because the spectral dimensions of the samples from source and target datasets are different, there are also the mapping modules to unify the dimensionality.

\subsection{Query Samples Mixing}


When given an input task consisting of a support set and a query set, either the support or query samples will be first processed by the mapping modules to transform their channel dimensionality. After that, each sample in the query set is mixed up with another randomly selected sample to generate the synthetic query samples. We choose to only mix up the query samples without support samples for the purpose of ensuring the correctness of the class prototypes caculated from these support samples. For two query samples $ x_i$ and $ x_k$ with the labels of $y_i$ and $y_k$, the mixing up process can be formally stated as Eq. \ref{Mix}.

\begin{equation}
	\begin{split}
	& \tilde{x} = \boldsymbol{M} \odot x_i + \boldsymbol{(1-M)} \odot x_k \\
	& \tilde{y}=\lambda_1 y_i + \lambda_2 y_k
	\end{split}
	\label{Mix}
\end{equation}

where $\boldsymbol{M}\in \left\{0,1\right\}^{W\times H}$ denotes a binary mask indicating where to drop out the pixels from the sample with the spatial size of $W\times H$. $\boldsymbol{1}$ is a binary mask filled with ones, and $\odot$ is the element-wise multiplication. $\tilde{x}$ is the synthetic query sample, and $\tilde{y}$ is the corresponding synthetic label, where $\lambda_1$ and $\lambda_2$ are the proportion of $y_i$ and $y_k$ in the synthetic label.

The proposed method follows CutMix \cite{37} to mix up the query samples. That is,  a randomly sampled patch in $x_i$ is removed and filled with the patch cropped from the same region of $x_k$. CutMix uses the proportion of the sampled patch to the entire sample as the value of $\lambda_1$ (i.e., $\lambda_2 = 1-\lambda_1$) to mix up the labels. Differently, we will caculate the values of $\lambda_1$ and $\lambda_2$ by using the attention maps returned from transformer, to take the different importance of the pixels into consideration. The details will be described in the following subsection of $E$.

\subsection{Transformer based Feature Extraction}

\begin{figure*}[ht]
	\centering
	\includegraphics[width=14.4cm,height=6.5cm]{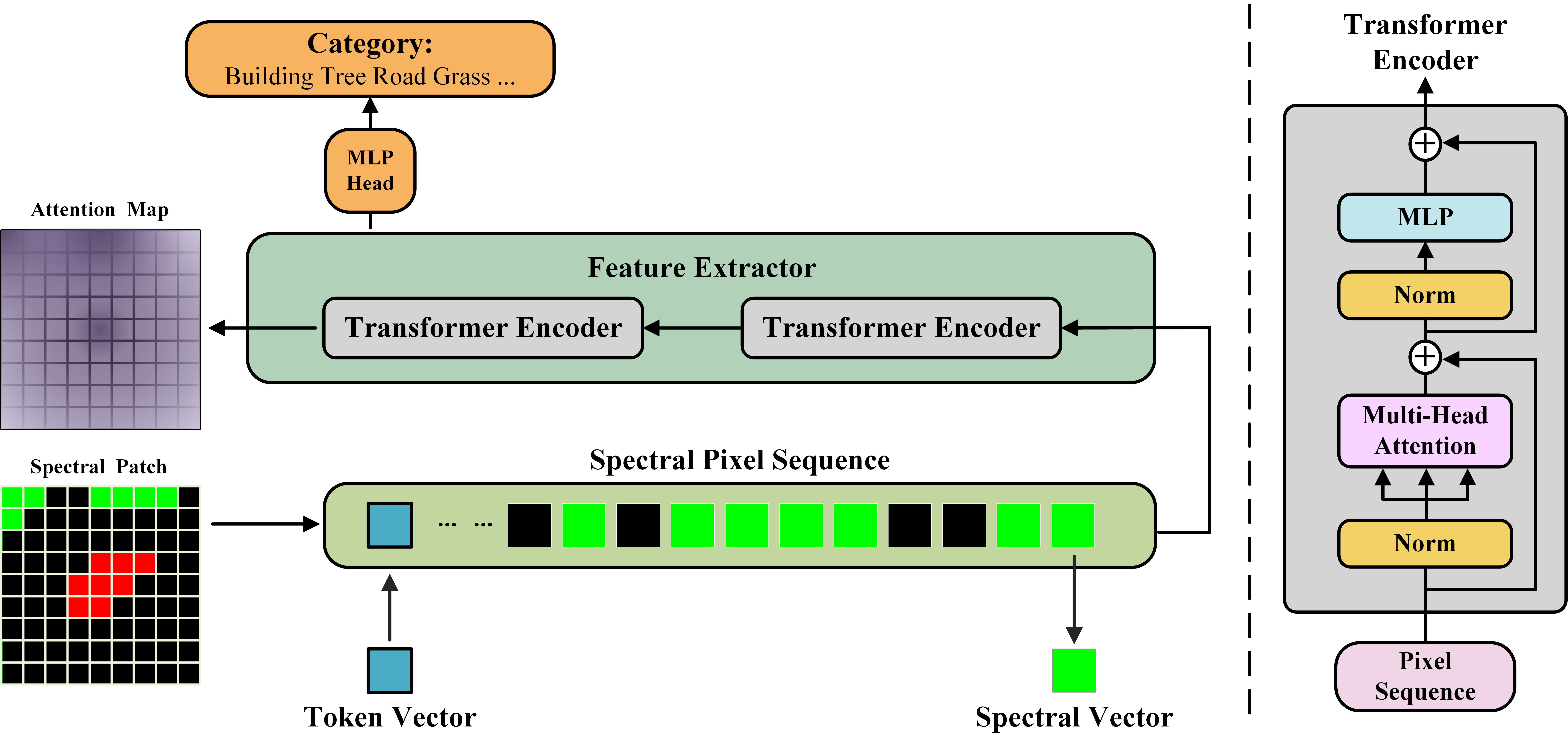}
	\caption{The transformer model for processing HSI patches, which outputs the spatial-spectral joint feature of each patch and the attentions reflecting the importance of different pixels in the patches.}
	\label{Network}
\end{figure*}

Once the synthetic query samples are generated, the original query samples, the synthetic query samples, and the support samples are input into the transformer module. The embedding features of the support samples and the synthetic query samples will be extracted. At the same time, the attention maps of the original query samples are also required. The whole process can be seen in Fig. \ref{Network}. 
 
When inputting the samples, i.e., the patches with the three-dimensional size of $W\times H\times D$, into transformer, each sample will be first pull into a sequence $x\in \mathbb{R}^{WH \times D}$ of which the composing items are the spectral vectors of the pixels. And then, the token vectors with $D$ dimension, which are used to learn global features of each sample, are randomly generated and appended into the sequences of each sample. After that, the features and the attention maps of each sample are obtained by directly inputting the updated sequences $x\in \mathbb{R}^{(WH+1) \times D}$ into transformer module.

A shown in Fig. \ref{Network}, the transformer module of APNT is composed of a set of stacked transformer encoders. Each encoder consists of a self-attention layer and a feedforward layer. For the self-attention layer, there exist three learnable projection matrixs $\boldsymbol{W}_q \in \mathbb{R}^{D \times D^{'}}$, $\boldsymbol{W}_k \in \mathbb{R}^{D \times D^{'}}$ and $\boldsymbol{W}_v \in \mathbb{R}^{D \times D^{'}}$. They will project the input sequences into different embedding spaces, as shown in Eq. \ref{Project}.

\begin{equation}
	\begin{split}
	& \boldsymbol{Q} = x \boldsymbol{W}_q \\
	& \boldsymbol{K} = x \boldsymbol{W}_k \\
	& \boldsymbol{V} = x \boldsymbol{W}_v \\
	\end{split}
	\label{Project}
\end{equation}

 where $\boldsymbol{Q}$, $\boldsymbol{K}$ and $\boldsymbol{V}$ are the projected matrices of the sequence $x$. With these projected matrices, the attention maps of each input sequence $a_x\in \mathbb{R}^{(WH+1) \times (WH+1)}$,  which capture the attention of each item in the sequence to the other items,  can be caculated as Eq. \ref{attention}. And the features of each item in the sequence are updated by Eq. \ref{feature}.
 
 \begin{equation}
 	\begin{split}
	&\boldsymbol{a_x} = \boldsymbol{Atttention(Q, K)}= softmax(\frac{\boldsymbol{Q}\boldsymbol{K}^T}{\sqrt{D^{'}}}) 
 	\end{split}
 	\label{attention}
 \end{equation}
 
  \begin{equation}
 \begin{split}
 & \boldsymbol{x^{'}}= \boldsymbol{Atttention(Q, K)}\boldsymbol{V}
 \end{split}
 \label{feature}
 \end{equation}

There are multiple heads of self-attention compution in transformer. The features returned from each head will be concatenated together and restored to the original dimension through a linear operation. The updated features will be further processed through the feedforward layer without changing the dimensions. That is, a new sequence of $ x^{''}\in \mathbb{R}^{(WH+1) \times D}$ will be produced by a transformer encoder for each sample. At the same time, the attention maps $a_x$ from different heads will be averaged. Through two transformer encoders, the updated taken vectors in the sequence $x^{''}$ generated by the last transformer encoder are selected as the feature vectors of the samples.  And the attention vectors capturing attentions of the samples to their constituent pixels will be also selected from the attention maps $a_x$ returned by the last transformer encoder. When reshaping the obtained attention vectors into the shape of $W \times H$, the required attention maps reflecting the importance of each pixel are obtained. 

\subsection{TransMix based Few-shot Learning Loss}

After obtaining the attention maps $\boldsymbol{A} \in \mathbb{R}^{W\times H}$ of these query samples,  the sum of the attention values belonging to the mixed region is directly used to derive the values of $\lambda_1$ and $\lambda_2$ for label mixing shown in Eq. \ref{Mix}. This is because that the sum of all values in a attention map $\boldsymbol{A}$ is 1. This derivation process can be experssed as Eq. \ref{tranmix}. 

 \begin{equation}
	\begin{split}
		& \lambda_1= \boldsymbol{M} \odot \boldsymbol{A_i} \\
		& \lambda_2= (\boldsymbol{1} - \boldsymbol{M}) \odot \boldsymbol{A_k} \\
	\end{split}
	\label{tranmix}
\end{equation}

where $\boldsymbol{A_i}$ and $\boldsymbol{A_k}$ represent the attention maps of the query smaple of $x_i$ and $x_k$. $\boldsymbol{M}$ represents the binary mask mentioned in Eq. \ref{Mix}. From Eq. \ref{tranmix}, it can be seen that the values of $\lambda_1$ and $\lambda_2$ represent the probability that the synthetic sample $\tilde{x}$ shown in Eq. \ref{Mix}  belongs to the classes of $y_i$ and $y_k$.

In addition, assuming the feature of each support smaple is $z_{i}$, the $c^{th}$ class prototype is computed as Eq.\ref{Prototype}.

\begin{equation}
	p_c= \frac{1}{|S^c|} \sum_{z_i\in {S^c}} z_i
	\label{Prototype}
\end{equation}

where $S^c$ represents the set of features of $c^{th}$ class support samples.   

Then, through predicting the synthetic query sample $\tilde{x}$ as the classes denoted by the labels of $y_i$ and $y_k$, the few-shot learning losses are caculated according to Eq. \ref{QLoss} and Eq. \ref{ALoss}. The $\tilde{z}_h$  in Eq. \ref{QLoss} and \ref{ALoss} represents the embedding feature of synthetic query sample $\tilde{x}_h$.   And  $p(\tilde{y}_h=y_i|\tilde{x}_h)$ means the probability that the model accurately predicts the synthetic query sample $\tilde{x}$ to the class of $y_i$. Meanwhile, $d(\tilde{z}_i, p_{y_i})$ denotes the  Euclidean distance between the feature of synthetic query sample $\tilde{x}_i$ and the class prototype of  $y_i$.

\begin{equation}
\begin{split}
\textit{L}^{i}_{fsl}=- \frac{1}{C\times M} \sum^{C\times M}_{h=1} log p(\tilde{y}_h=y_i| \tilde{x}_h) \\
where \quad p(\tilde{y}_h=y_i| \tilde{x}_h) = \frac{exp( d(\tilde{z}_h, p_{y_i}))}{\sum^{C_s}_{j=1}exp(d(\tilde{z}_h, p_j))}
\end{split}
\label{QLoss}
\end{equation}

\begin{equation}
	\begin{split}
		\textit{L}^{k}_{fsl}=- \frac{1}{C\times M} \sum^{C\times M}_{z=1} log p(\tilde{y}_h =y_k| \tilde{x}_h) \\
		where \quad p(\tilde{y}_h =y_k | \tilde{x}_h) = \frac{exp( d(\tilde{z}_h, p_{y_k}))}{\sum^{C_s}_{j=1}exp(d(\tilde{z}_h, p_j))}
	\end{split}
	\label{ALoss}
\end{equation}

Because the synthetic query samples only partially belong to the classes of the samples used for mixing, 
the total loss used for updating the model is caculated as Eq. \ref{totalloss}, which uses the values of $\lambda_1$ and $\lambda_2$ to balance the losses of $L^i_{fsl}$ and $L^k_{fsl}$. 

\begin{equation}	
	L = \lambda_1 L^i_{fsl} + \lambda_2 L^k_{fsl}
	\label{totalloss}
\end{equation}


\section{Experiments}
In order to validate the effectiveness of the proposed method, extensive experiments have been done. In this section, we describe the experimental setup and results.

\subsection{Experimental Setup}

\subsubsection*{Dataset}

In order to fairly evaluate the performance of the proposed method, the datasets which had been widely used in related works were selected for training and testing in our work. We used the Indian Pine, University of Pavia, and Salinas datasets as the target datasets, and used the Chikusei dataset as the source dataset. We give a brief introduction about these datasets.

\begin{figure}[htbp]
	\centering
	\subfigure[]{
		\includegraphics[width=2.5cm,height=3.5cm]{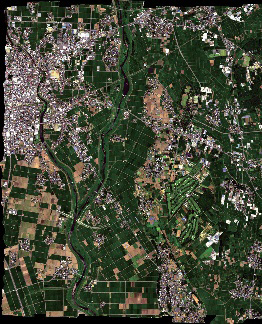}
	}
	\subfigure[]{
		\includegraphics[width=2.5cm,height=3.5cm]{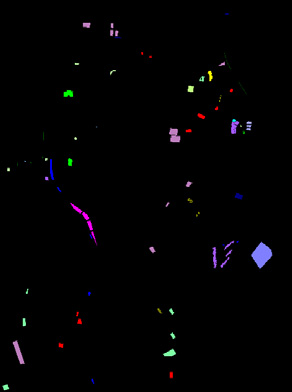}
	}
	\subfigure[]{
		\includegraphics[width=2.5cm,height=3.5cm]{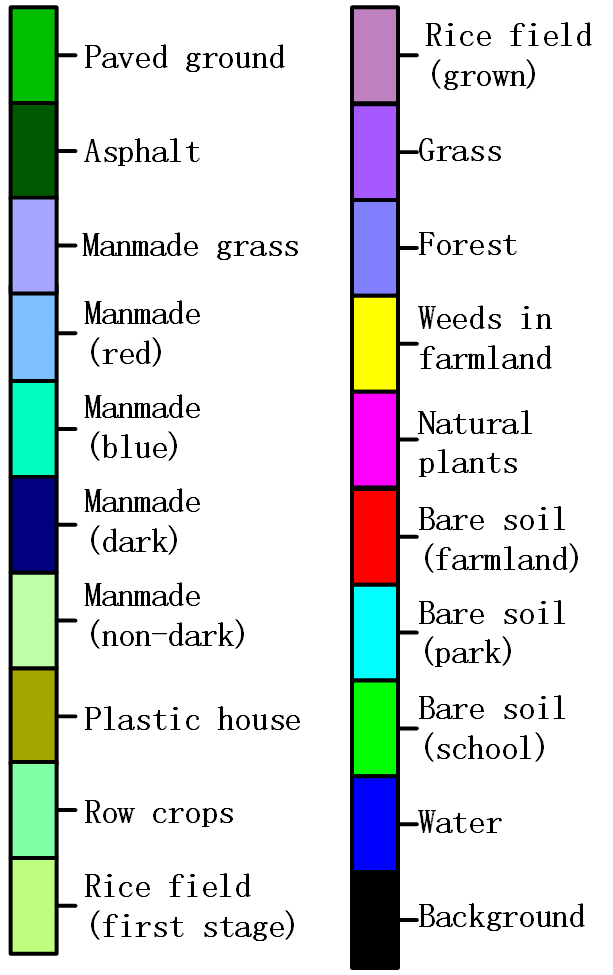}
	}
	
	\caption{The color map, label map and label color of Chikusei dataset}
	\label{fig1}
\end{figure}

(a) \textbf{Chikusei}: It was taken by Hyperspec-VNIR-C sensor in Chikusei, Japan. The dataset contains 128 bands with the wavelength ranging from $343nm$ to $1018 nm$, and $ 2517 \times 2335$ pixels with a spatial resolution of $2.5 m$. As shown in Fig. \ref{fig1}, there are 19 classes of land cover, including urban and rural areas. 

\begin{figure}[htbp]
	\centering
	\subfigure[]{
		\includegraphics[width=2.5cm,height=2.5cm]{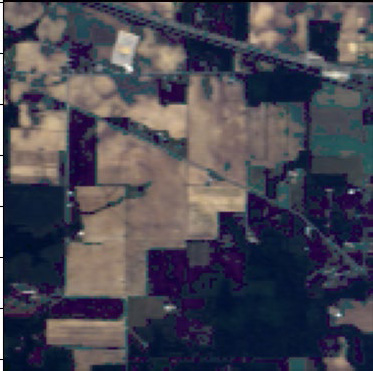}
	}
	\subfigure[]{
		\includegraphics[width=2.5cm,height=2.5cm]{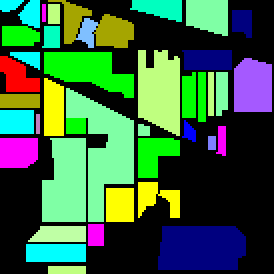}
	}
	\subfigure[]{
		\includegraphics[width=2.5cm,height=2.5cm]{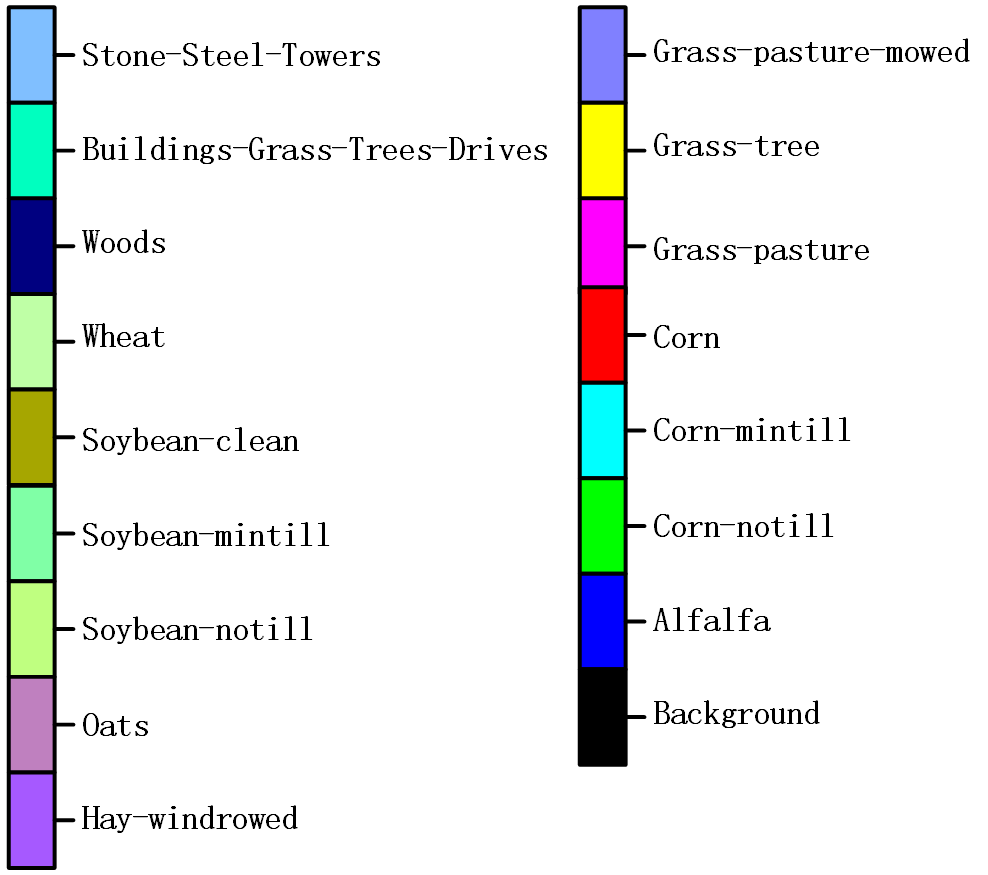}
	}
	
	\caption{The  color map, label map and label color of IP dataset}
	\label{fig2}
\end{figure}

(b) \textbf{Indian Pines (IP)}: It was imaged by an airborne visible infrared imaging spectrometer in Indiana, USA. This dataset contains 200 bands with the wavelength ranging from $0.4-2.5(10^{-6})m $. The size is $145\times145$ pixels, with a spatial resolution of about $ 20m$. As shown in Fig. \ref{fig2}, there are 16 classes of land cover  including crops and natural vegetation.

\begin{figure}[htbp]
	\centering
	\subfigure[]{
		\includegraphics[width=1.5cm,height=3cm]{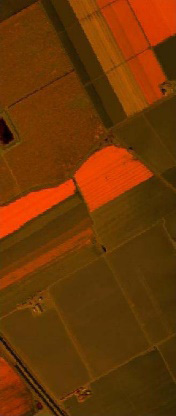}
	}
	\subfigure[]{
		\includegraphics[width=1.5cm,height=3cm]{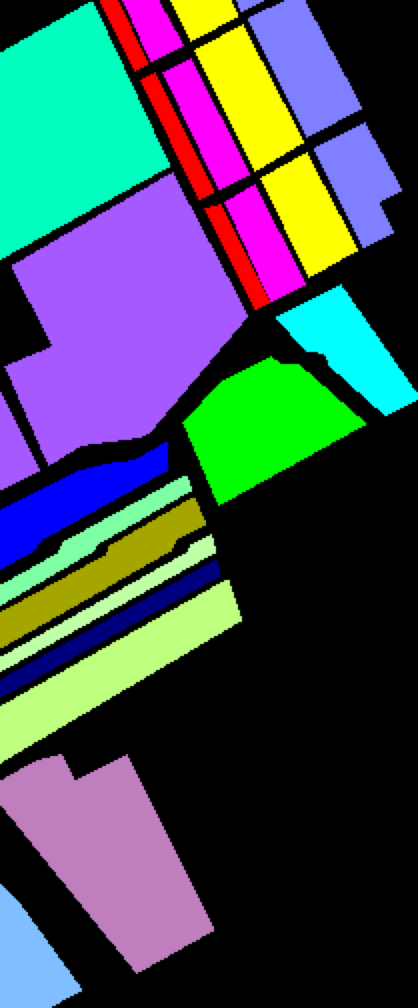}
	}
	\subfigure[]{
		\includegraphics[width=3.5cm,height=3cm]{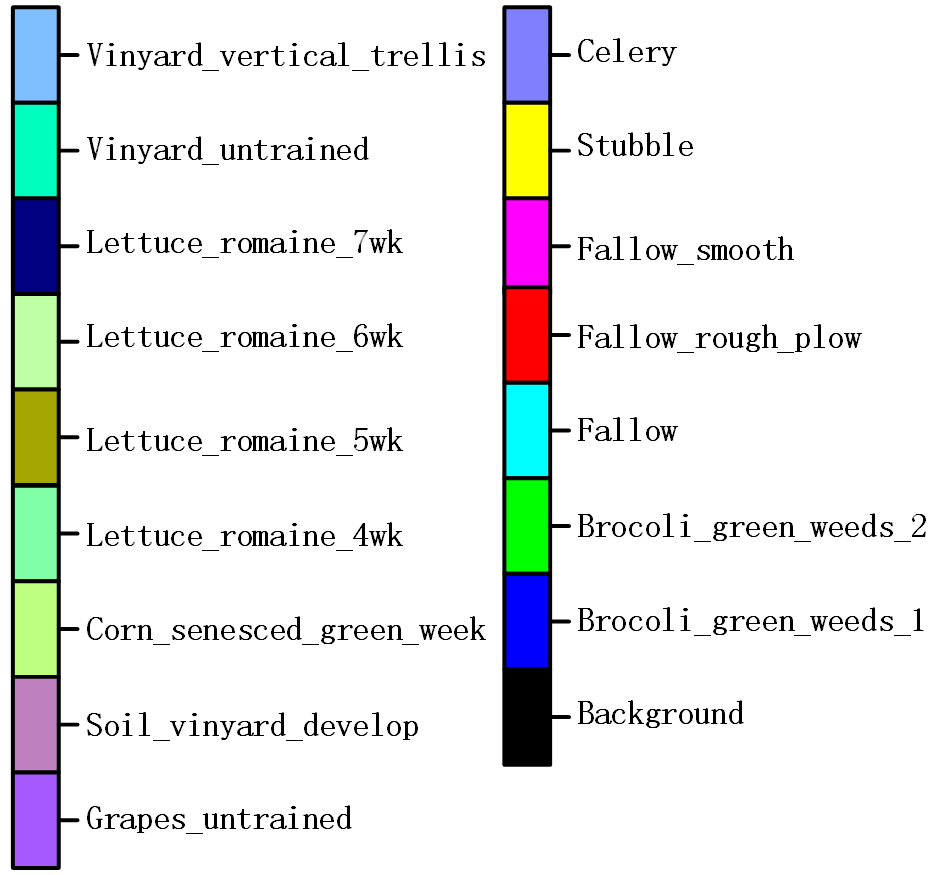}
	}
	
	\caption{ The color map, label map and label color of SA dataset}
	\label{fig3}
\end{figure}

(c) \textbf{Salinas (SA)}: This dataset was also captured  by the airborne visible infrared imaging spectrometer in the Salinas Valley, California, USA. It contains 204 wavebands with a size of $512 \times 217$ pixels, and a spatial resolution of about $3.7m$. As shown in Fig. \ref{fig3}, there are 16 classes of land cover, which include  vegetables, exposed soil, etc. 

\begin{figure}[htbp]
	\centering
	\subfigure[]{
		\includegraphics[width=2cm,height=3cm]{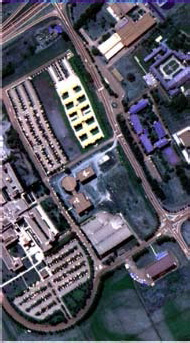}
	}
	\subfigure[]{
		\includegraphics[width=2cm,height=3cm]{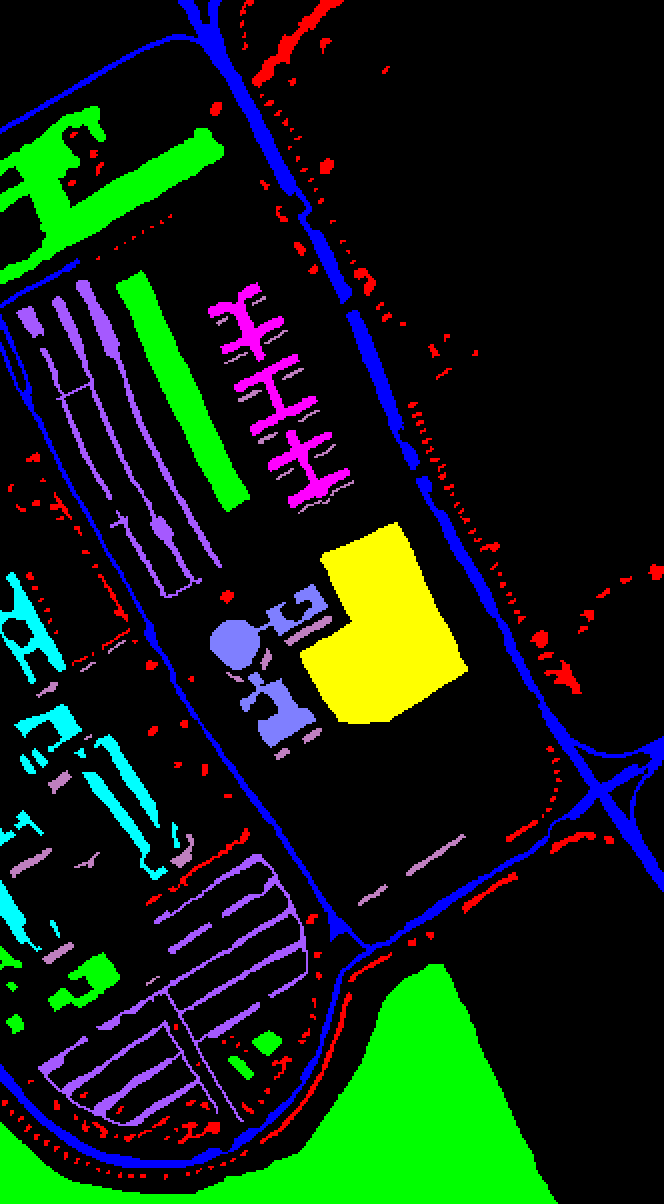}
	}
	\subfigure[]{
		\includegraphics[width=1.2cm,height=3cm]{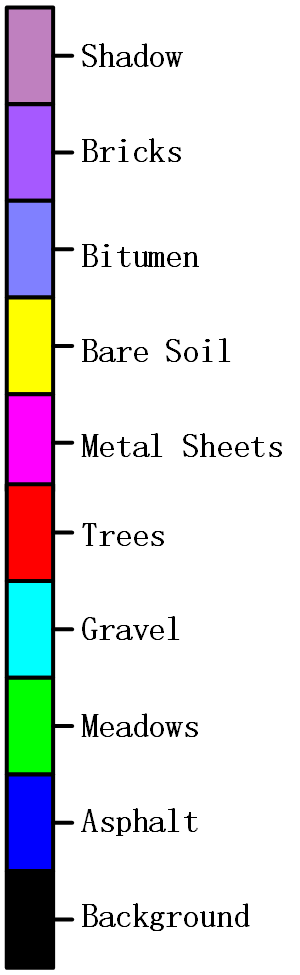}
	}
	
	\caption{The  color map, label map and label color of PU dataset}
	\label{fig4}
\end{figure}

(d) \textbf{Pavia University (PU)}: This dataset was imaged by the German airborne reflective optical spectral imager in the city of Pavia, Italy. The dataset contains 103 wavebands, with a size of $610 \times 340$ pixels and a spatial resolution of about $1.3m$. As shown in Fig. \ref{fig4}, there are 9 classes of land cover, including trees, asphalt roads, bricks, etc. 

\subsubsection*{Metric}
Also for the fairness of comparison, we adopted the metrics of overall accuracy rate (OA), average accuracy rate (AA) and kappa coefficient which have been widely used to evaluate the performance of proposed method. The definitions of these metrics can be referred to the work of \cite{30}.

\subsubsection*{Implementation and Configuration}
 For the proposed method, the mapping modules are implemented with a Conv2D layer with $1 \times 1 $ kernel. The feature extractor is implemented by two transformer encoders, each of which includes a self-attention layer and a fully connected feedforward layer. The specific parameters related to the implementation are shown in the Table \ref{model-modules}, where Head\_Dim denotes the number of attention channels, and Feed\_Dim represents the hidden layer dimension of the feedforward layer.

All linear and convolutional layers were normalized using Xavier, and the Adam optimizer was used to optimize  the model, with an initial learning rate of 0.001. All comparison experiments used the same 10 random seeds to fairly compare the mean of AA, OA, and kappa. The patches with the size of $9\times9$ pixels were cut from HSIs. Moreover, as mentioned early,  there are five samples per class selected from the target dataset to fine-tune the model trained on the source dataset. And they are augmented to 200 samples per class through cropping and resizing, which is consistent with the work of \cite{33}.  There are 3000 training iterations adopted in our experiments. Among them, only the samples from source datasets are used in the first 1000 training iterations, and the remaining 2000 iterations are run on these augmented samples from target datasets.

\begin{table}[h]
	\small
	\caption{The architecture of APNT model}
	\label{model-modules}
	\centering
	\renewcommand{\arraystretch}{1.5}
	\setlength{\tabcolsep}{2.7mm}{
		\begin{tabular}{c|c||c}
			\hline
			\multicolumn{2}{c||}{Model Architecture} & SIZE \\ \hline\hline
			\multicolumn{2}{c||}{INPUT} & 9$\times$9$\times d$ \\ \hline\hline
			\multirow{1}{*}{MAPPING} & Conv2d & 1$\times$1(100) \\  \hline
			\multirow{3}{*}{TRANSFORMER ENCODER1} 
			& Head & 8 \\
			& Head\_Dim & 64 \\         
			& Feed\_Dim & 1024 \\ \hline
			\multirow{3}{*}{TRANSFORMER ENCODER2} 
			& Head & 8 \\
			& Head\_Dim & 64 \\         
			& Feed\_Dim & 1024 \\ \hline
			\multirow{2}{*}{OUTPUT} & Feature & 1$\times$160 \\
			& Attention Map & 9$\times$9 \\ \hline\hline
		\end{tabular}
	}
\end{table}


\subsection{Comparison with Related Methods}
To verify the performance of the proposed method, we selected several typical classification methods for comparison, including DFSL\cite{24}, DCFSL\cite{29}, CMFSL\cite{31}, Gia-CFSL\cite{32}, RPCL\cite{33},  MRLF\cite{34}, and HFSL\cite{35}.  In the experiment, according to the specific training process of each model, DFSL only used these augmented samples from target dataset for training. DCFSL, CMFSL, Gia-CFSL, MRLF, RPCL and our proposed method used both the samples from source datasets and target datasets, while HFSL used the natural image dataset of Mini-ImageNet for pre-training and was fine-tuned on these augmented samples from target datasets. In addition, in order to prove that the proposed method still had good performance without pre-training on source datasets, the performance of APNT* method which  is only trained on these augmented samples from target datasets is also evaluated.


\begin{table*}[h]
	\small
	\caption{Comparison results on IP dataset (5 labeled samples per class)}
	\label{IP-result}
	\centering
	\renewcommand{\arraystretch}{1.5}
	\setlength{\tabcolsep}{2mm}{
		\begin{tabular}{|c|c|c|c|c|c|c|c|c|c|}
			\hline			
			\diagbox{Class}{Method} & \textbf{DFSL}   & \textbf{DCFSL} & \textbf{CMFSL} &  \textbf{Gia-CFSL} & \textbf{MRLF} & \textbf{RPCL} & \textbf{HFSL} & \textbf{APNT*} & \textbf{APNT}  \\ \hline
			1& 96.75&	96.34 & 96.34 & 92.44 &	98.05 &	99.51  & 99.51 & \textbf{100.0} & \textbf{100.0} \\  \hline
			2& 36.38&	48.10 & 48.70 & 43.08 &	51.78 & 63.53  & 55.06 & \textbf{69.09} & 68.67 \\  \hline
			3& 38.34&	53.27 & 59.96 & 51.24 &	55.43 &	\textbf{64.85} & 61.38 & 63.61  & 62.15 \\  \hline
			4& 77.16&	81.59 & 78.88 & 75.60 &	77.28 &	90.99  & 88.15 & 91.81 & \textbf{92.16}\\  \hline
			5& 73.92&	73.77 & 77.13 & 70.98 &	\textbf{80.61} & \textbf{80.61}  & 74.62 & 78.56 & 79.12 \\  \hline
			6& 86.25&	86.14 & 80.36 & 83.08 & 91.26 & 89.85 & 79.26 & 91.02 & \textbf{91.70}\\  \hline
			7& 97.10&	99.57 & \textbf{100.00} & 99.57 & 98.7  & \textbf{100.0} & 99.57 & \textbf{100.0}  & \textbf{100.0} \\  \hline
			8& 81.82&	82.98 & 90.53 & 85.48 &	83.07 &	90.13  & \textbf{98.50} & 97.80 &	97.53\\  \hline
			9& 75.56&\textbf{100.0} & \textbf{100.0} & \textbf{100.0} &	98.67 &	\textbf{100.0} & 98.67 & 98.67 & \textbf{100.0}\\  \hline
			10& 52.22&	63.45 & 63.52 & 64.38 &	66.61 & 68.61  & 63.06 & 70.92 & \textbf{73.22}\\ \hline
			11& 59.96&	60.09 & 60.65 & 63.96 &	63.53 &	65.81 & 64.04 & 64.88 & \textbf{66.11}\\  \hline
			12& 36.56&	44.73 & 48.16 & 50.49 &	53.91 & \textbf{64.54} & 64.35 & 60.00 & 61.48\\ \hline
			13& 98.00&	99.00 & \textbf{99.35} & 97.95 &	99.05 & 97.15 & 99.05 & 97.00& 97.25\\ \hline
			14& 84.63&	81.72 & 81.33 & 81.94 &	83.76 & 89.37 & 90.27 & \textbf{90.62} & 90.21\\ \hline
			15& 74.10&	71.44 & 72.31 & 64.51 &	82.10 & 85.88 & 91.15 & \textbf{91.76} & 89.97\\ \hline
			16& \textbf{100.0}	&	98.86 & 99.66 & 99.32 & 97.50& 98.18 & 96.59 & 96.36 & 95.91 \\  \hline
			OA& 61.69&  65.85 & 66.85 & 65.70 & 69.46 & 74.55 & 71.74  & 75.68 & \textbf{76.06}\\ 
			AA& 73.05 & 77.57 & 78.56 & 76.50 & 80.24 & 84.31 & 82.79 & 85.13  & \textbf{85.34}\\ 
			Kappa &56.78  &  61.59   & 62.67 & 61.28 & 65.69 & 71.34 & 68.18 & 72.68  & \textbf{73.10}\\  \hline
		\end{tabular}
	}
\end{table*}

\begin{figure*}[htbp]
	\centering
	\subfigure[]{
		\includegraphics[width=3cm,height=3cm]{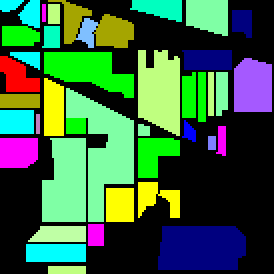}
	}
	\quad
	\subfigure[]{
		\includegraphics[width=3cm,height=3cm]{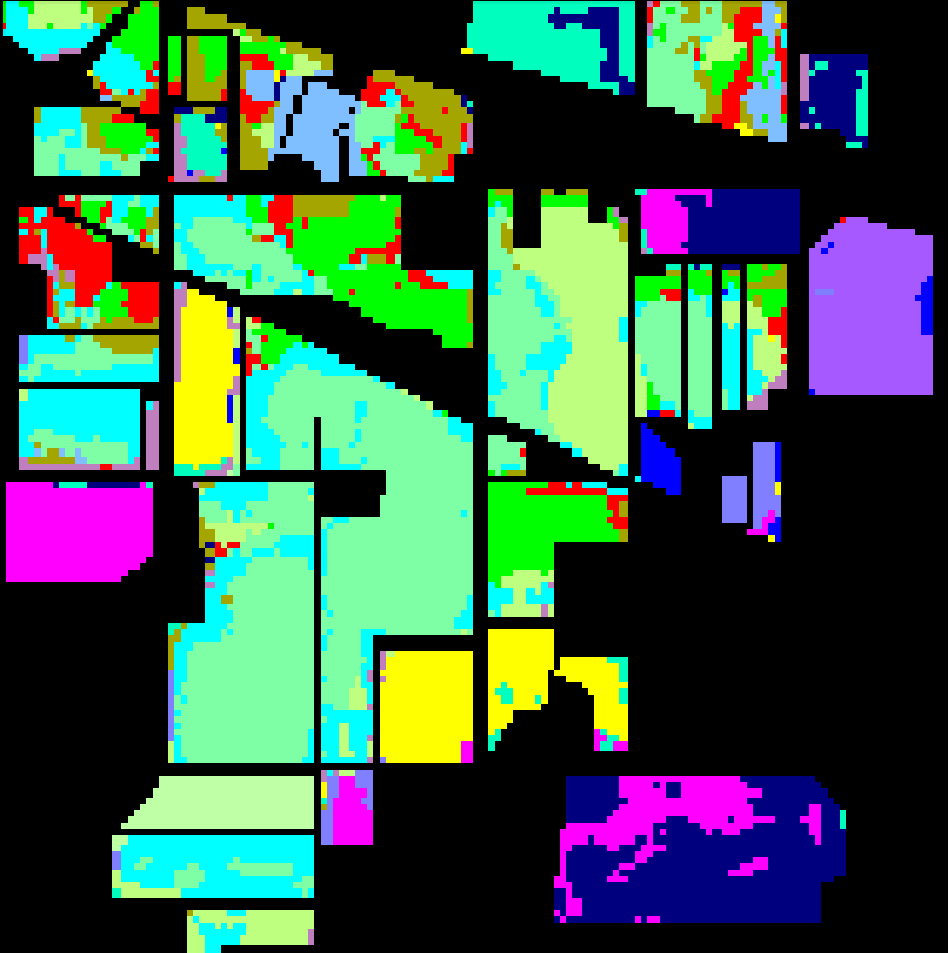}
	}
	\quad
	\subfigure[]{
		\includegraphics[width=3cm,height=3cm]{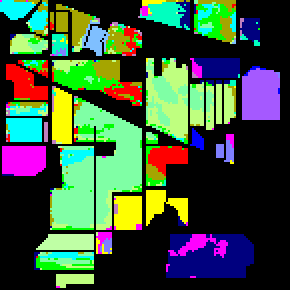}
	}
	\quad
	\subfigure[]{
		\includegraphics[width=3cm,height=3cm]{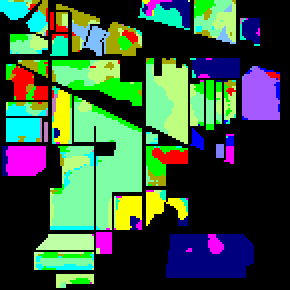}
	}
	\quad
	\subfigure[]{
		\includegraphics[width=3cm,height=3cm]{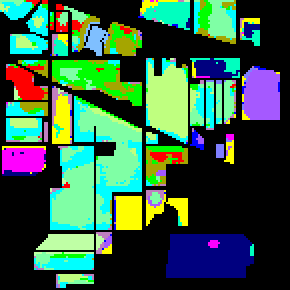}
	}
	\quad
	\subfigure[]{
		\includegraphics[width=3cm,height=3cm]{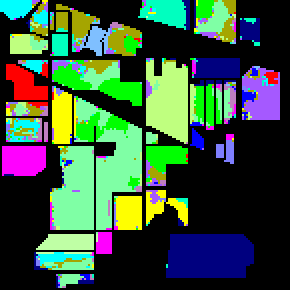}
	}
	\quad
	\subfigure[]{
		\includegraphics[width=3cm,height=3cm]{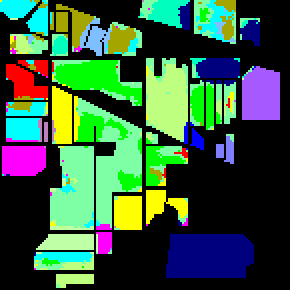}
	}
	\quad
	\subfigure[]{
		\includegraphics[width=3cm,height=3cm]{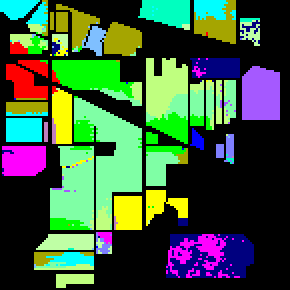}
	}
	\quad
	\subfigure[]{
		\includegraphics[width=3cm,height=3cm]{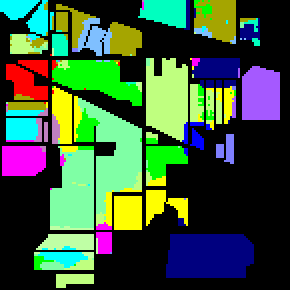}
	}
	\quad
	\subfigure[]{
		\includegraphics[width=3cm,height=3cm]{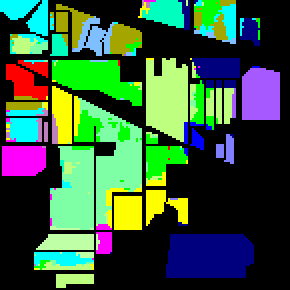}
	}
	
	\caption{ Data visualization and classification diagrams of target scenes obtained using different methods on IP dataset,  include: (a) ground reality map, (b) DFSL (61.69\%), (c) DCFSL (65.85\%), (d) CMFSL (66.85\%),  (e) Gia-CFSL (65.70\%), (f) MRLF (69.46\%), (g) RPCL (74.55\%), (h) HFSL (71.74\%), (i) APNT* (75.68\%), and (j) APNT (76.06\%).}
	\label{fig5}
\end{figure*}

\begin{table*}[h]
	\small
	\caption{Comparison results on PU dataset (5 labeled samples per class)}
	\label{PU-result}
	\centering
	\renewcommand{\arraystretch}{1.5}
	\setlength{\tabcolsep}{2mm}{
		\begin{tabular}{|c|c|c|c|c|c|c|c|c|c|}
			\hline
			\diagbox{Class}{Method} & \textbf{DFSL}   & \textbf{DCFSL} & \textbf{CMFSL} &  \textbf{Gia-CFSL} & \textbf{MRLF} & \textbf{RPCL} & \textbf{HFSL} & \textbf{APNT*} & \textbf{APNT}  \\ \hline
			1& 73.43&	79.07 & 81.50 & 78.23 &	78.77 & 82.41 & 75.64 &90.55 & \textbf{91.41}\\ \hline
			2& \textbf{89.25}&	85.43 & 83.45 & 87.73 &	83.54 & 78.65 & 88.62 & 86.22 &	87.58\\ \hline
			3&	48.09&	62.82 & 68.72 & 63.13 &	68.83 & 67.76  & 82.35 & 81.44 & \textbf{82.60}\\ \hline
			4&	84.72&	92.69 & 90.14 & 91.41 &	93.11 & 93.66 & 93.08 & \textbf{93.93} & 93.18\\ \hline
			5&	99.65&	99.47 & 99.71 & 99.24 & 99.71 & 99.45 & \textbf{99.96} & 98.94 & 99.41\\ \hline
			6&	67.81&	74.82 & 76.04 & 72.36 &	79.72 & 78.88 & 84.29 & \textbf{87.06} &84.79\\ \hline
			7&	64.48&	75.12 & 80.96 & 76.83 &	89.26 & 81.51 & 85.43 & 98.41 & \textbf{99.10}\\ \hline
			8&	67.37&	65.22 & 77.57 & 70.85 &	73.39 & 86.42  & \textbf{95.26} & 85.40 & 87.94\\ \hline
			9&	92.92&	\textbf{98.45} & 98.13 & 98.27 &	96.36 & 95.55 & 97.91 & 89.76 &	90.66\\ \hline
			OA&  79.63&  81.28 & 82.01 & 82.31 &   82.41  & 81.58  & 87.14 & 88.09 & \textbf{88.83}\\ 
			AA&76.41& 81.46 & 82.29 & 84.02 &   84.74  & 84.92 & 89.17 & 90.19 & \textbf{90.74}\\ 
			Kappa&73.05  & 75.74   & 77.05 & 76.92 &77.40  & 76.58 & 83.27 & 84.60 & \textbf{85.55}\\ \hline
		\end{tabular}
	}
\end{table*}

\begin{figure*}[htbp]
	\centering
	\subfigure[]{
		\includegraphics[width=2.7cm,height=4.5cm]{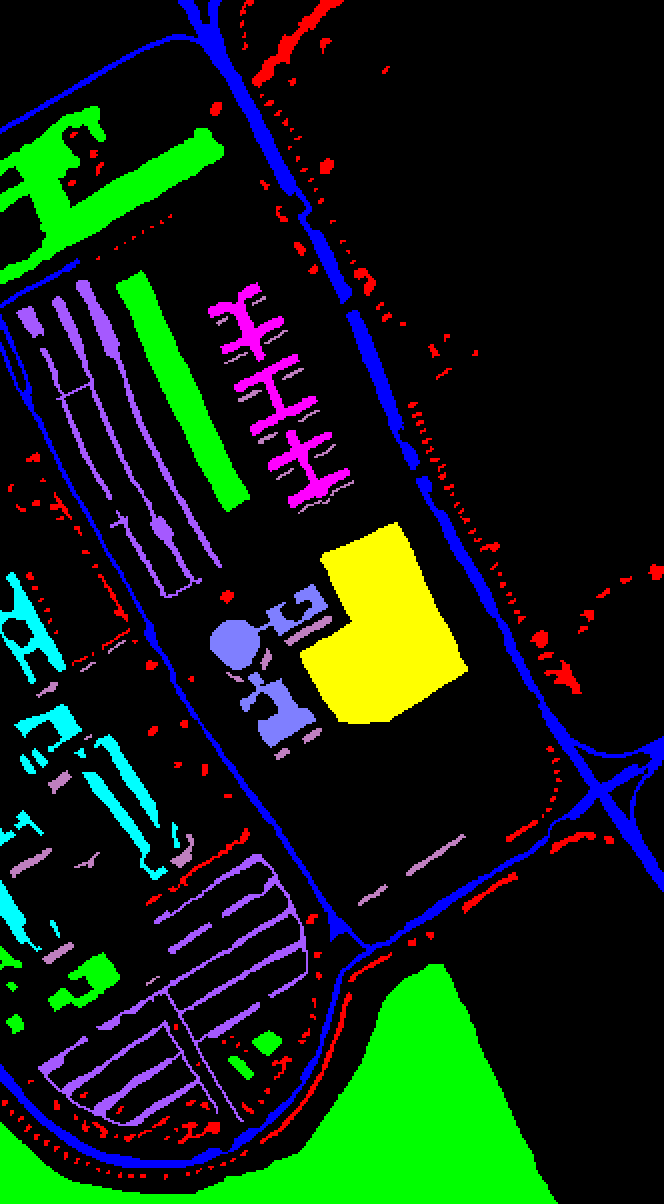}
	}
	\quad
	\subfigure[]{
		\includegraphics[width=2.7cm,height=4.5cm]{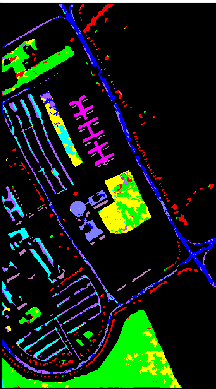}
	}
	\quad
	\subfigure[]{
		\includegraphics[width=2.7cm,height=4.5cm]{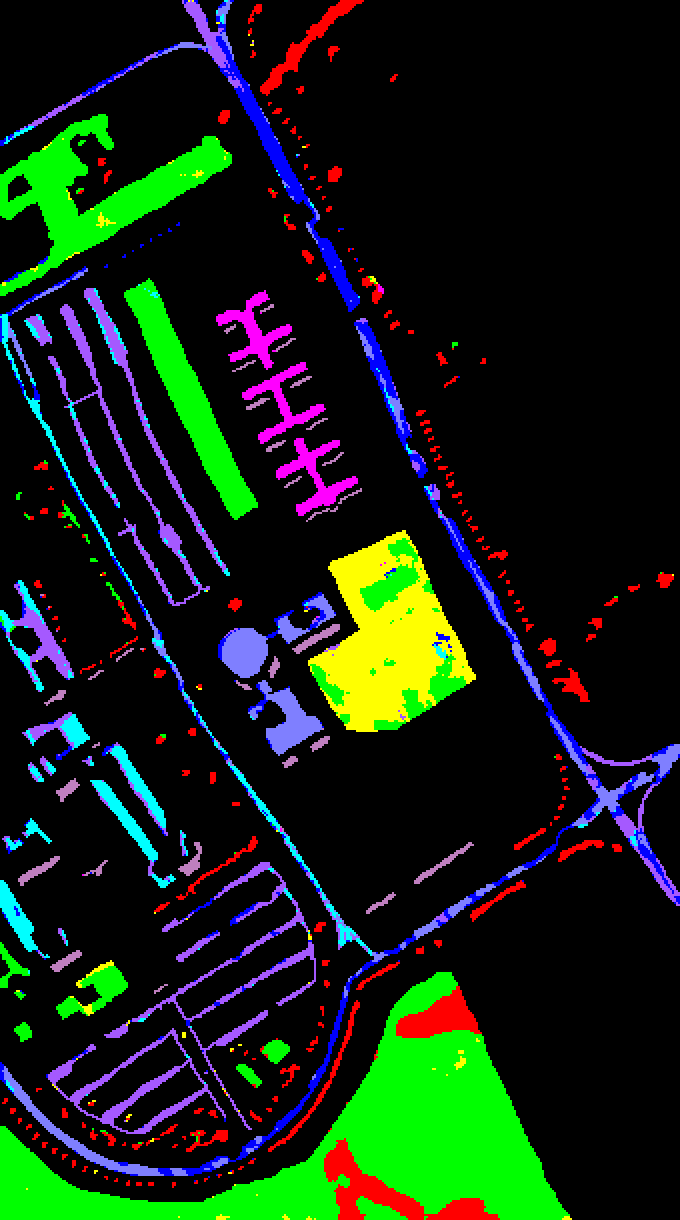}
	}
	\quad
	\subfigure[]{
		\includegraphics[width=2.7cm,height=4.5cm]{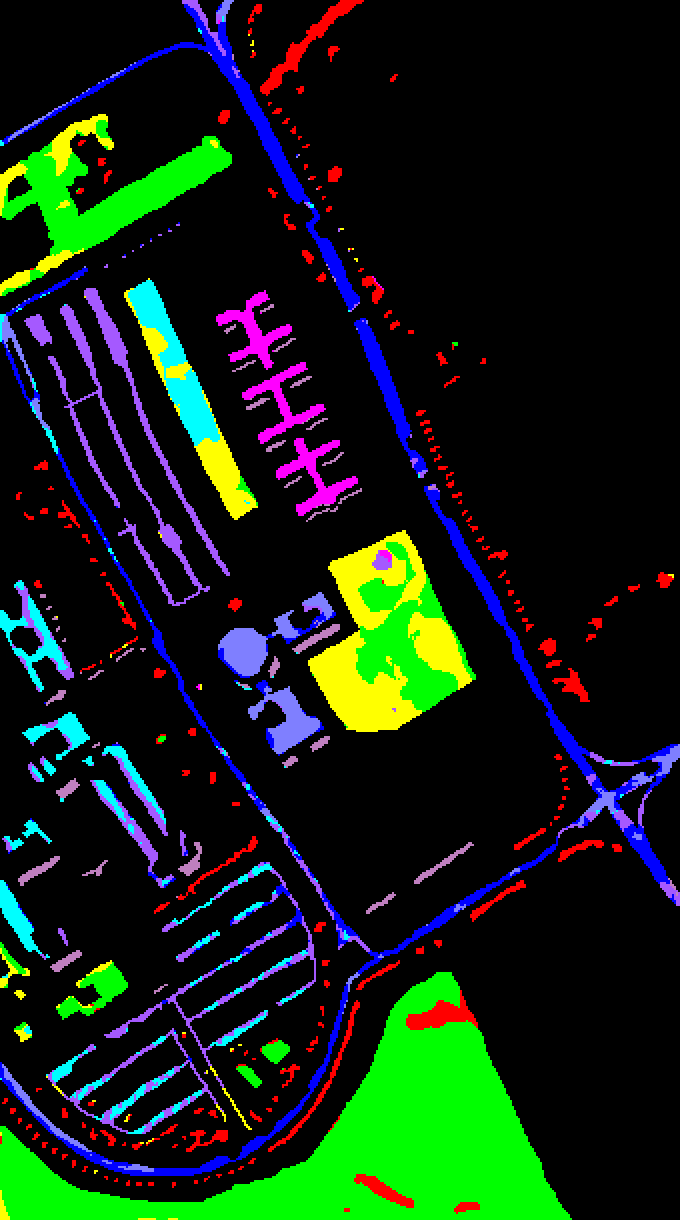}
	}
	\quad
	\subfigure[]{
		\includegraphics[width=2.7cm,height=4.5cm]{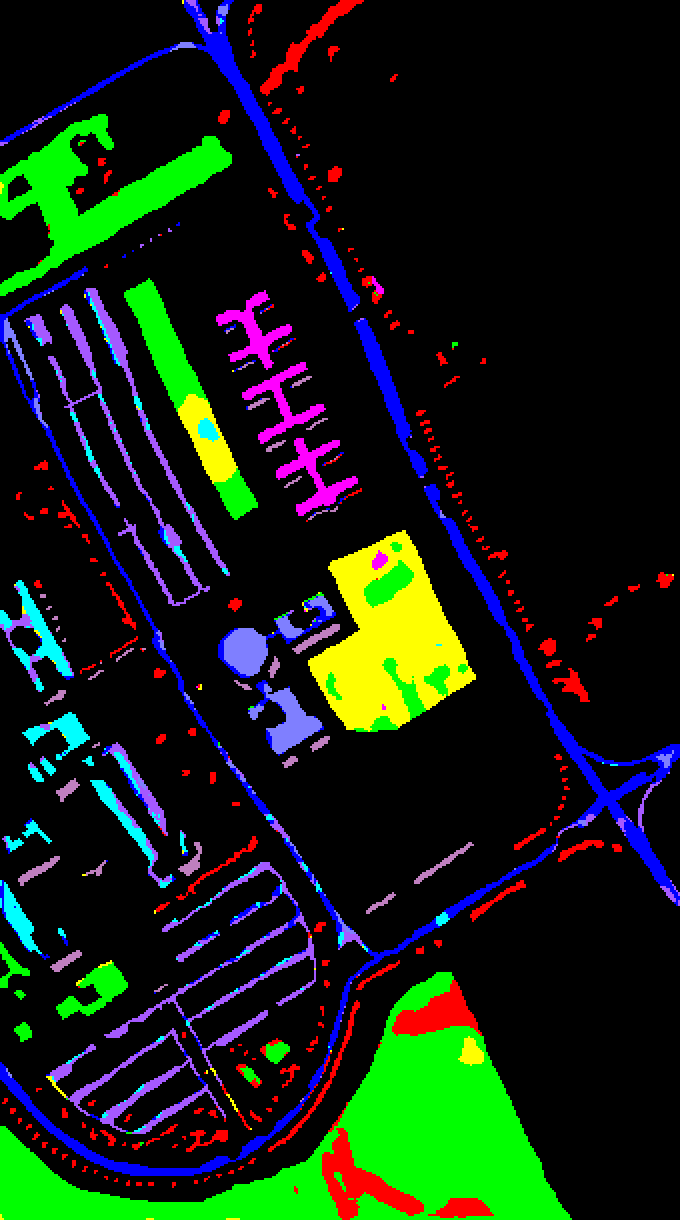}
	}
	\quad
	\subfigure[]{
		\includegraphics[width=2.7cm,height=4.5cm]{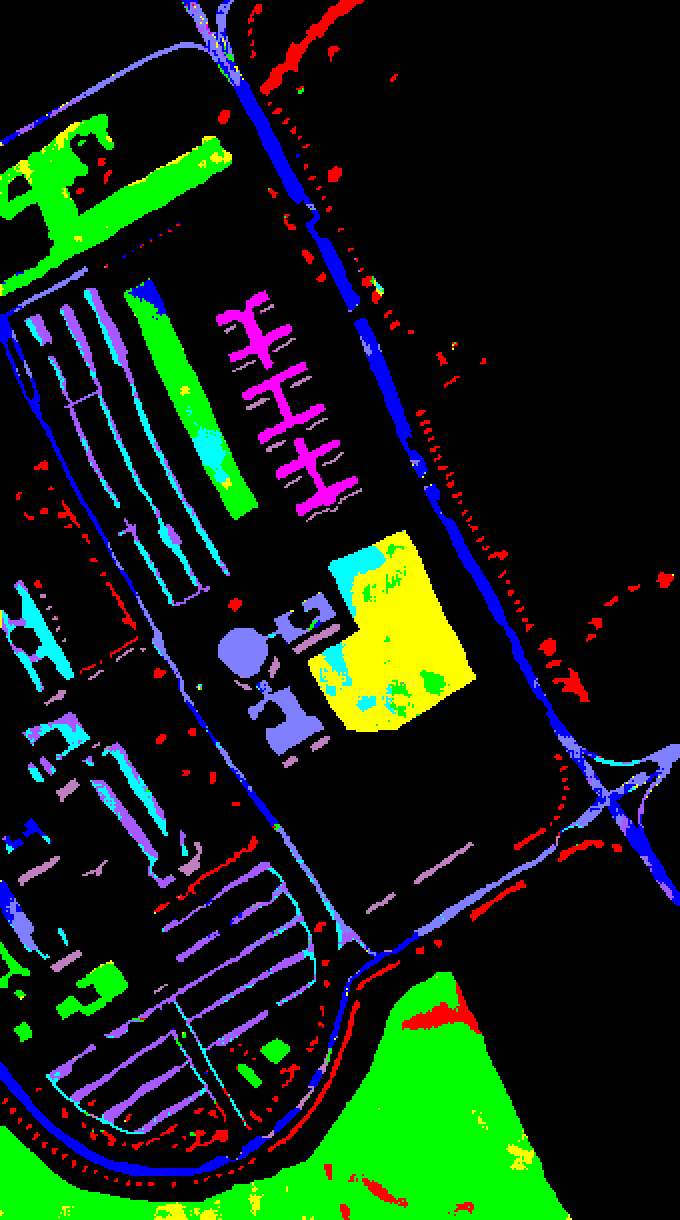}
	}
	\quad
	\subfigure[]{
		\includegraphics[width=2.7cm,height=4.5cm]{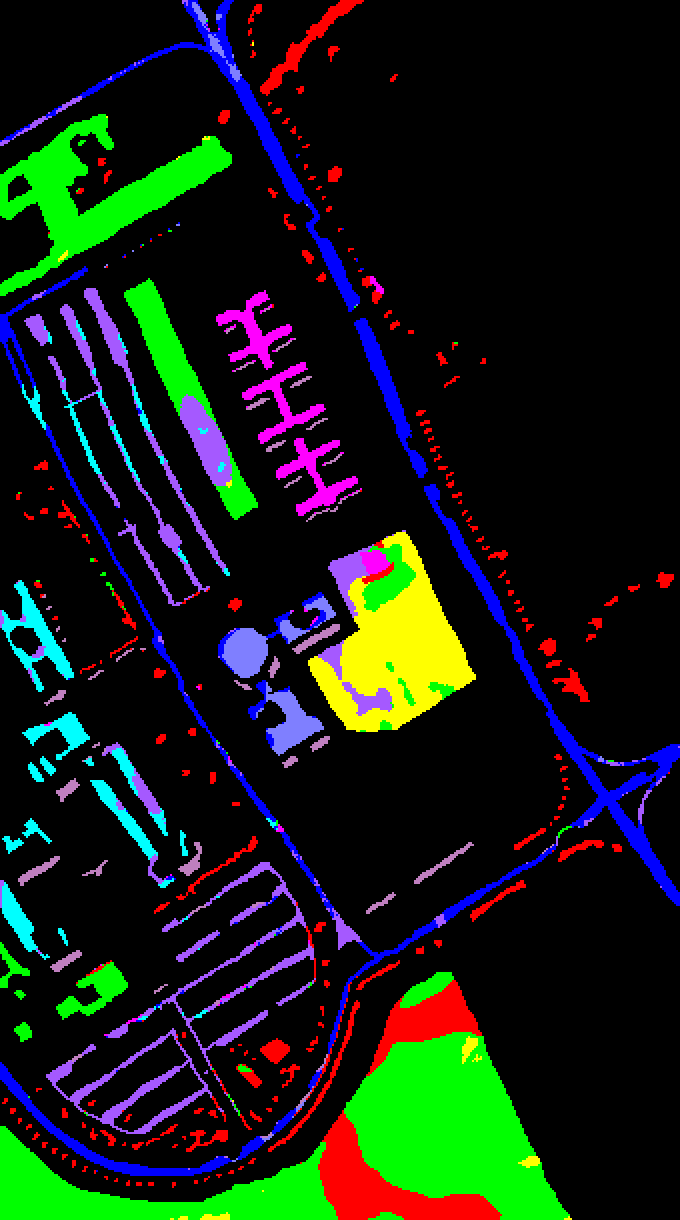}
	}
	\quad
	\subfigure[]{
		\includegraphics[width=2.7cm,height=4.5cm]{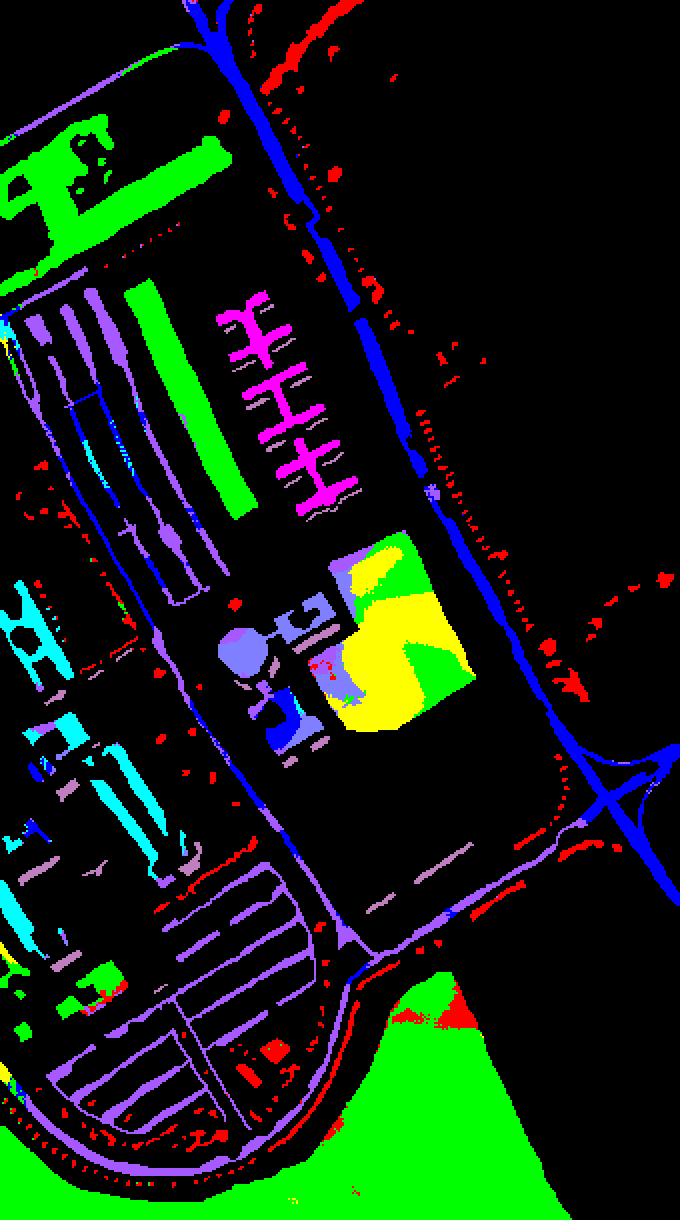}
	}
	\quad
	\subfigure[]{
		\includegraphics[width=2.7cm,height=4.5cm]{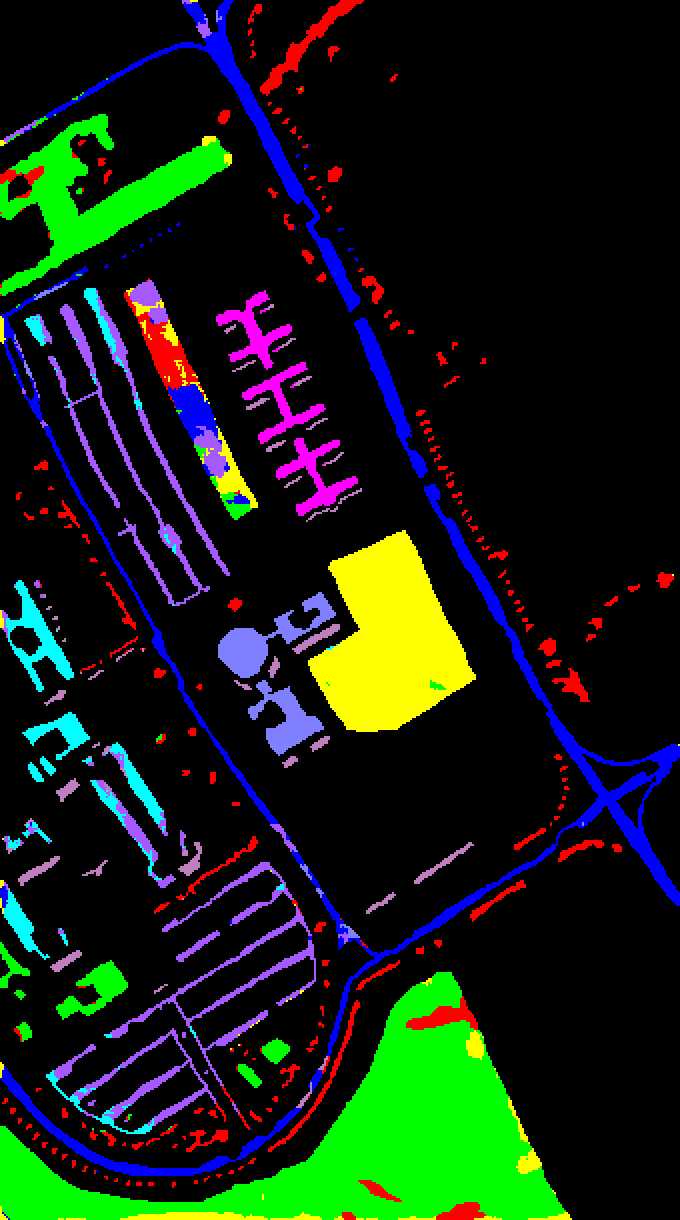}
	}
	\quad
	\subfigure[]{
		\includegraphics[width=2.7cm,height=4.5cm]{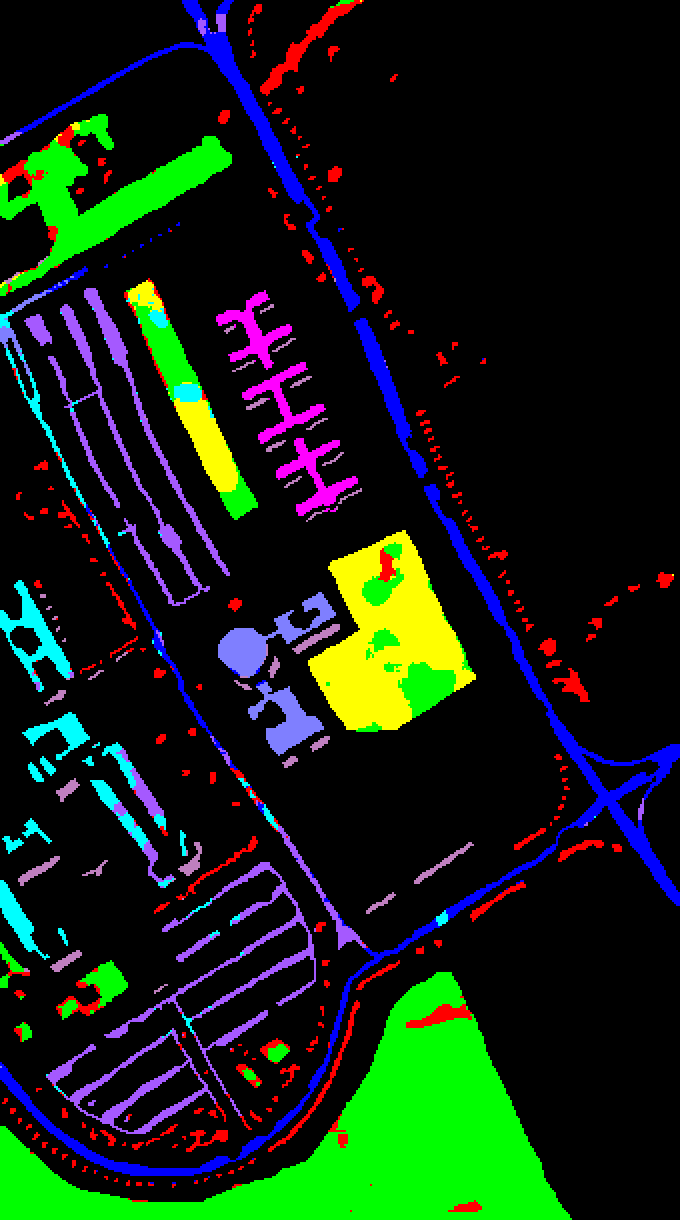}
	}
	
	\caption{ Data visualization and classification diagrams of target scenes obtained using different methods on PU dataset, including: (a) ground reality map, (b) DFSL (79.63\%), (c) DCFSL (81.28\%), (d) CMFSL (82.01\%), (e) Gia-CFSL (82.31\%), (f) MRLF (82.41\%), (g) RPCL (81.58\%), (h) HFSL (87.14\%), (i) APNT* (88.09\%), and (j) APNT (88.83\%).}
	\label{fig6}
\end{figure*}

\begin{table*}[h]
	\small
	\caption{Comparison results on SA dataset (5 labeled samples per class)}
	\label{SA-result}
	\centering
	\renewcommand{\arraystretch}{1.5}
	\setlength{\tabcolsep}{2mm}{
		\begin{tabular}{|c|c|c|c|c|c|c|c|c|c|}
			\hline
			\diagbox{Class}{Method} & \textbf{DFSL}   & \textbf{DCFSL} & \textbf{CMFSL} &  \textbf{Gia-CFSL} & \textbf{MRLF} & \textbf{RPCL} & \textbf{HFSL} & \textbf{APNT*} & \textbf{APNT}  \\ \hline
			1 &	73.92  & 99.27  & 97.13 & 99.06 &	\textbf{99.65}  & 99.53 & 98.84 & 99.39 & 99.32\\ \hline
			2 &	96.85 &	98.89  & 99.16 & 99.34 &	99.97  & 99.00  & 92.47 & 99.98 & \textbf{100.00}\\ \hline
			3 &	96.28 &	93.21  & 90.72 & 89.88 &	89.37  & 90.90 & 96.15 & \textbf{96.91} &	95.27\\ \hline
			4 &	99.11 &	99.52  & 99.17 & 98.76 & 99.47 & 99.40 & \textbf{99.60}  & 99.45 &	99.58\\ \hline
			5 &	80.72 &	91.26  & 93.03 & 89.16 &	91.18  & 91.40 & \textbf{95.79} & 92.58 &	94.77\\ \hline
			6 & 91.63 & 99.38  & 99.46 & 97.63 & 99.93  & 97.92 & \textbf{98.83}  & 96.11 &	95.59\\ \hline
			7 &	97.73 &	\textbf{99.40}  & 97.73 & 99.29 &	99.06  & 98.93 & 95.25 & 97.91  &	97.67\\ \hline
			8 & 82.33 &	75.83  & 70.05 & 76.80 &	79.85  & \textbf{83.62} & 72.06 & 81.66  &	78.84\\ \hline
			9 &	94.44 &	98.70  & 99.20 & 98.32 & 99.90  & 99.85  & 99.23 & 99.91 &	\textbf{99.98}\\ \hline
			10&	80.96 &	84.91  & 84.27 & 81.31 &	87.17  & 86.72 & \textbf{91.37} & 91.26  & 90.64\\ \hline
			11&	93.38 &	98.14  & 96.69 & 96.43 &	98.67  & 98.25 & 98.42 & 99.01 & \textbf{99.44}\\ \hline
			12&	97.94 &	\textbf{99.78}  & 98.36 & 98.76 &	99.27  & 98.95& 98.90 & 98.09  &	97.01\\ \hline
			13&	95.79 &	99.15  & \textbf{99.67} & 98.11 &	98.40  & 99.13 & 99.01 & 99.07  &	99.08 \\ \hline
			14&	\textbf{98.87} &	98.59  & 98.81 & 97.89 &	98.78  & 98.10 & 96.80 & 94.88 & 94.61\\ \hline
			15&	71.13 &74.36  & 77.11 & 74.89 & 78.08  & 76.63 & 82.29 & 81.31 & \textbf{84.31}\\ \hline
			16&	90.57 &	89.51  & 87.08 & 82.14 &	92.5  & 91.81 & 91.51 & \textbf{98.76} & 97.51\\ \hline
			OA&  86.95 & 89.14 & 87.99 & 88.45 &   90.79 & 91.15 & 89.43  & \textbf{91.94}  & 91.65 \\ 
			AA& 90.08  &  93.74   & 92.98 & 92.36 &94.77 & 94.39 & 94.16  & \textbf{95.39} & 95.23\\ 
			Kappa&85.51 &  87.94   & 86.67 & 87.16 &89.69  & 90.15 & 88.28  & \textbf{91.04} & 90.72\\ \hline
		\end{tabular}
	}
\end{table*}

\begin{figure*}[htbp]
	\centering
	\subfigure[]{
		\includegraphics[width=2.7cm,height=4.5cm]{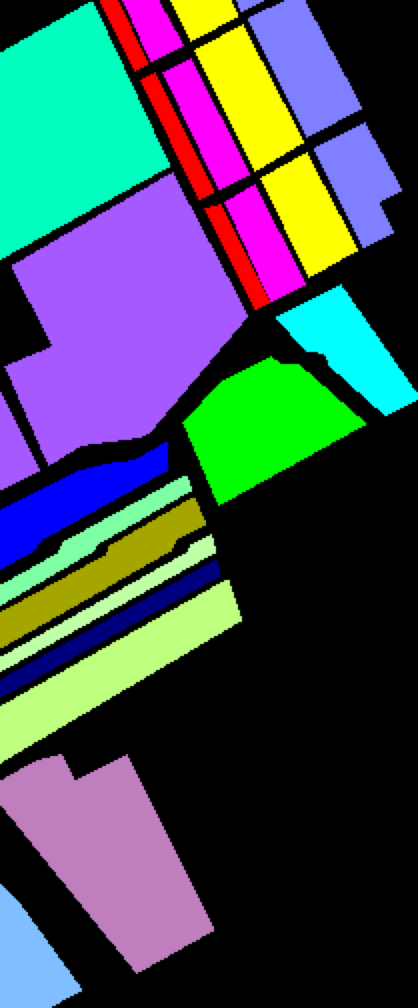}
	}
	\quad
	\subfigure[]{
		\includegraphics[width=2.7cm,height=4.5cm]{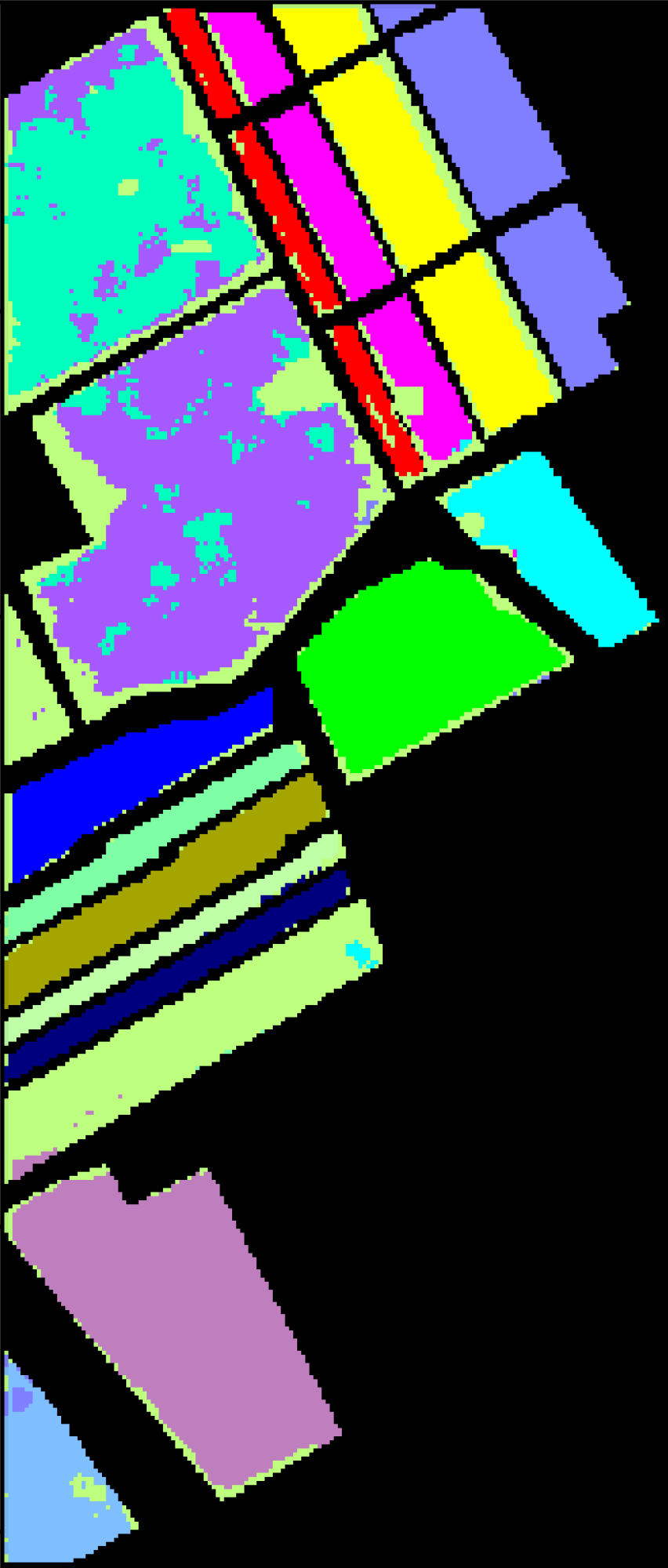}
	}
	\quad
	\subfigure[]{
		\includegraphics[width=2.7cm,height=4.5cm]{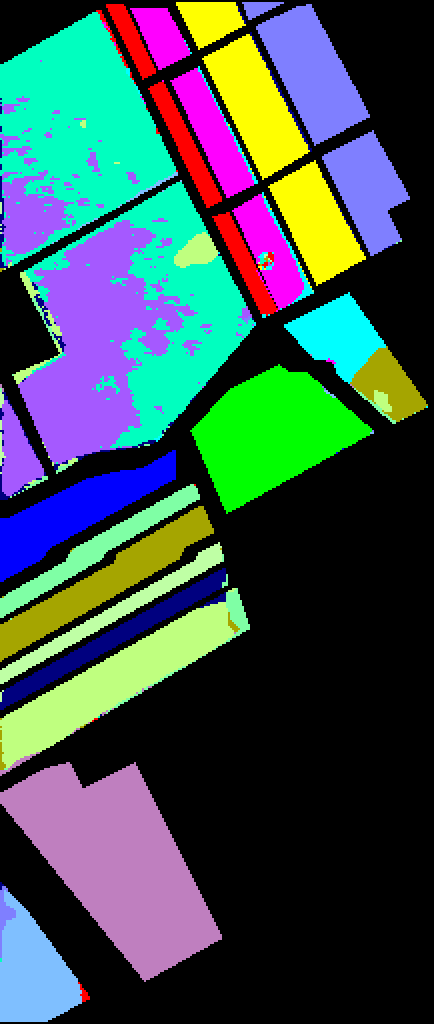}
	}
	\quad
	\subfigure[]{
		\includegraphics[width=2.7cm,height=4.5cm]{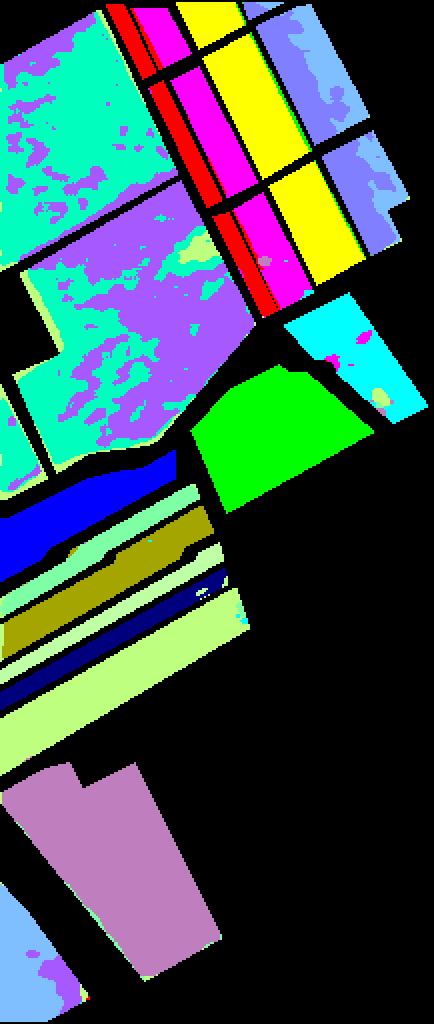}
	}
	\quad
	\subfigure[]{
		\includegraphics[width=2.7cm,height=4.5cm]{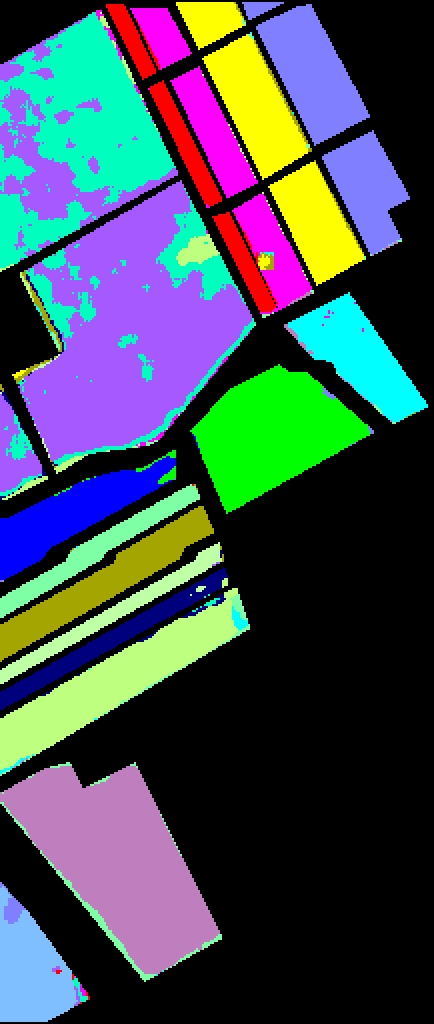}
	}
	\quad
	\subfigure[]{
		\includegraphics[width=2.7cm,height=4.5cm]{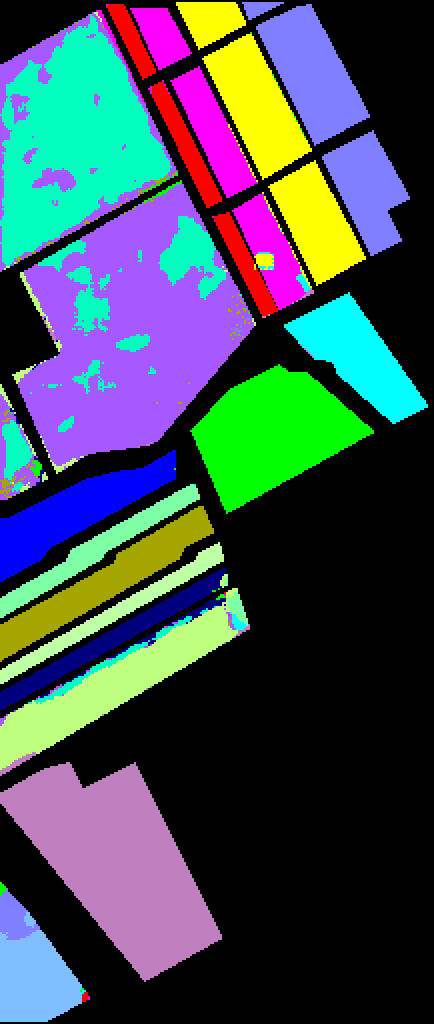}
	}
	\quad
	\subfigure[]{
		\includegraphics[width=2.7cm,height=4.5cm]{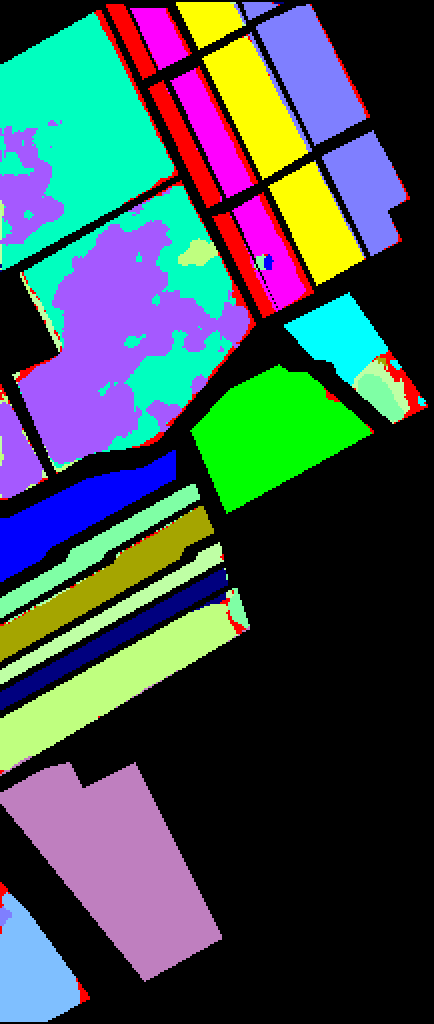}
	}
	\quad
	\subfigure[]{
		\includegraphics[width=2.7cm,height=4.5cm]{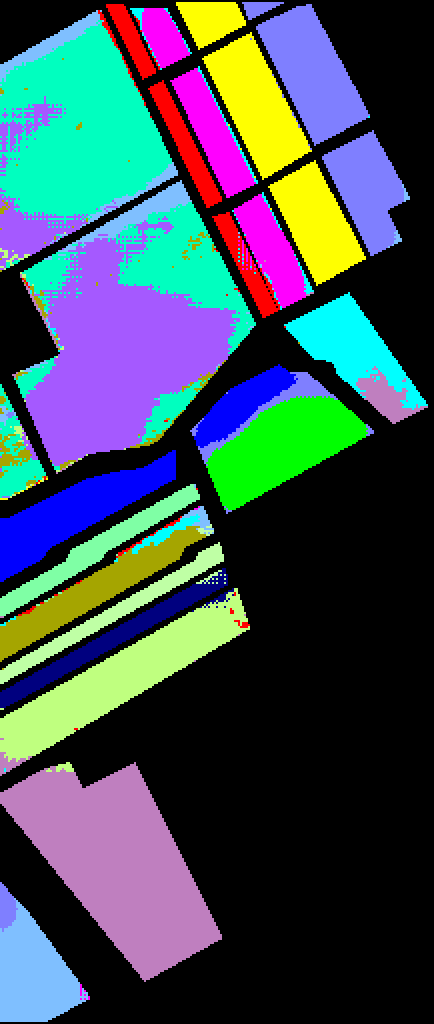}
	}
	\quad
	\subfigure[]{
		\includegraphics[width=2.7cm,height=4.5cm]{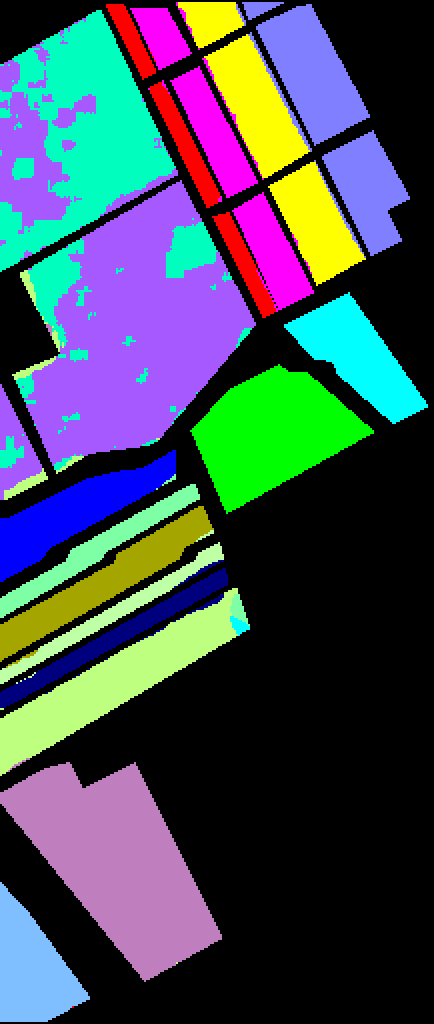}
	}
	\quad
	\subfigure[]{
		\includegraphics[width=2.7cm,height=4.5cm]{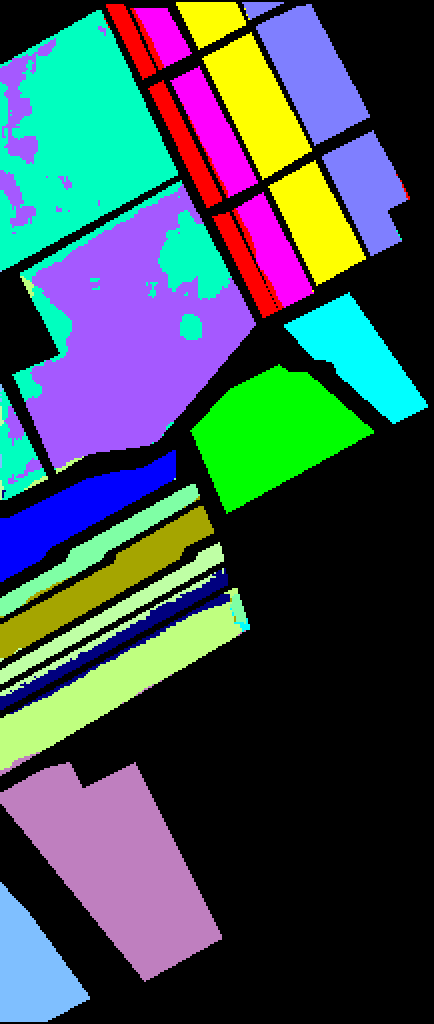}
	}
	
	\caption{ Data visualization and classification diagrams of target scenarios obtained using different methods on SA dataset, including: (a) Ground Realistic Map, (b) DFSL (86.95\%), (c) DCFSL (89.14\%), (d) CMFSL (87.99\%), (e) Gia-CFSL (88.45\%), (f) MRLF (90.79\%), (g) RPCL (91.15\%), (h) HFSL (89.43\%), (g) APNT* (91.94\%), and (h) APNT (91.65\%).}
	\label{fig7}
\end{figure*}

Table \ref{IP-result} to Table \ref{SA-result} report the specific classification results of each class, as well as the AA, OA, and kappa values of each method on the three target datasets. From the comparison results, it can be seen that APNT* and APNT have superior classification results compared to other methods. Among them, APNT outperforms the other results by $1.5\%$, $1.7\%$, and $0.5\%$ on the IP, PU, and SA datasets, respectively, reaching the optimal results on class 1, 4, 6, 7, 9, 10, and 11 of IP dataset, class 1, 3, and 7 of PU dataset, and class 2, 9, 11, and 15 of SA dataset. This result demonstrates the effectiveness of the proposed method. In the meanwhile, APNT* outperforms the other models by $1.1\%$, $0.9\%$, and $0.8\%$ on the IP, PU, and SA datasets respectively, reaching the optimal results on class 1, 2, 7, 14, and 15 of IP dataset, class 4 and 6 of PU dataset, and class 3 and 16 of SA dataset. Because APNT* does not use the sampes from source dataset for pre-training, this result shows that the proposed method can eliminate the dependence on existing large-scale datasets and can achieve comparative performance by only using these few-shot labeled samples in target tasks. This will facilitate the use of the proposed method in a range of applications.

The classification maps corresponding to each method are shown in Figures \ref{fig5} to \ref{fig7}. In these maps, colored pixels are labeled, and black pixels are unlabeled and displayed as the background. By observing the results displayed in the classification maps, it can be observed that the classification maps generated by the proposed method are most similar to the real ground map. In particular, we can notice the reduction of misclassified boundary points. These results also confirm the effective improvement of the classification results by the proposed method.

\subsection{Evaluation on Boundary Patches}
To validate the improvement of the predicting on boundary pixels, we processed the datasets and extracted all the boundary patches for testing. In the hyperspectral datasets, when all the pixels in the patches don't belong to the same class, we define such patches as boundary patches. Generally speaking, these boundary patches are the hard samples, and it is more difficult to predict their classes. 

\begin{figure}[htbp]
	\centering
	\includegraphics[width=8.8cm,height=6.4cm]{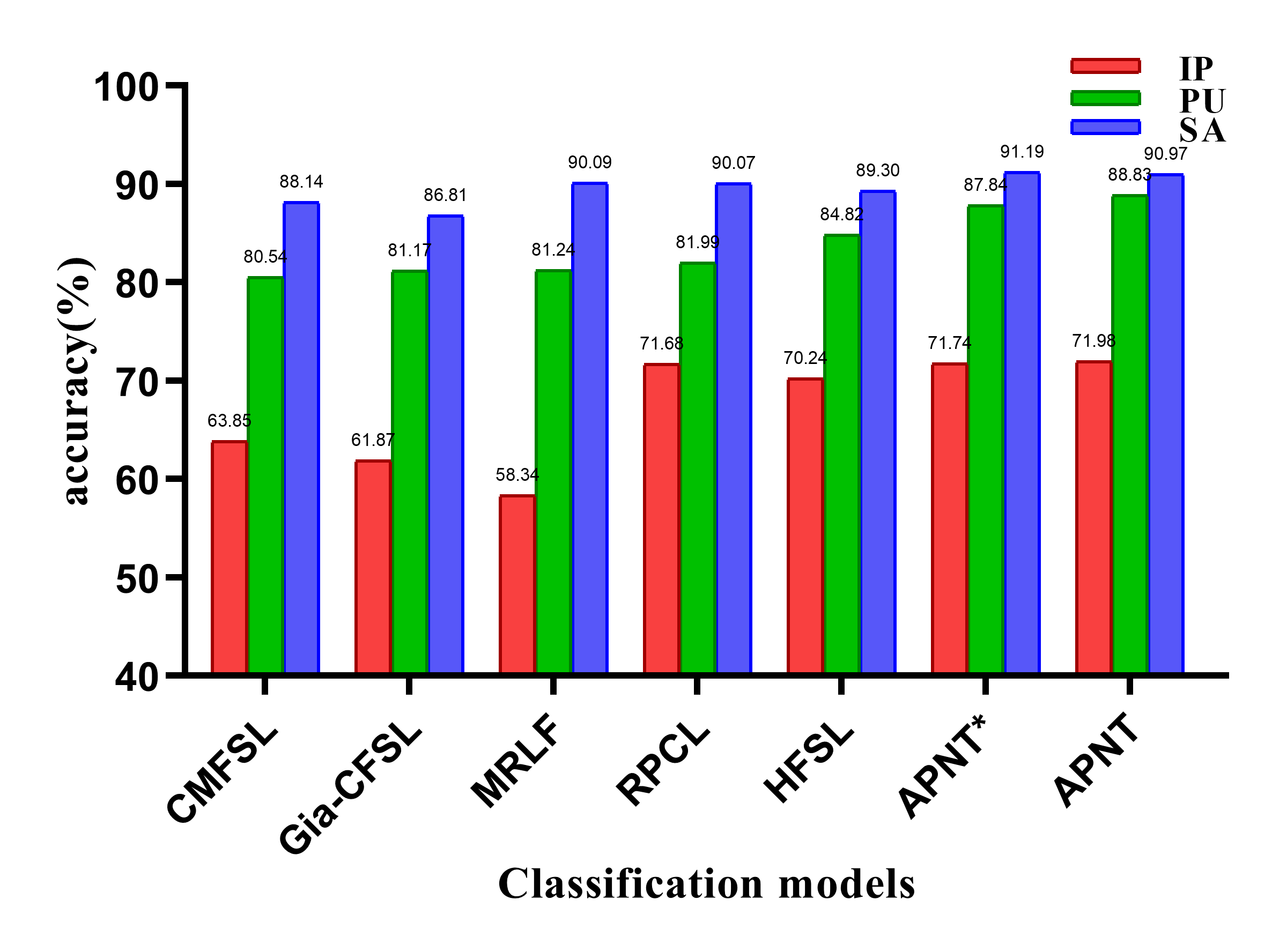}
	\caption{Comparison results of boundary point on IP, SA, PU datasets(5 labeled samples)}
	\label{Edge-result}
\end{figure}

We  calculated the overall accuracy of various methods on  all the boundary patches, which are shown in Fig. \ref{Edge-result}. As can be seen from the results, consistent with the assumption, APNT has the highest overall prediction performance on the boundary patches over three datasets. 

\subsection{Ablation Experiments}

While applying transformer to pay different attention to different pixels and extract better features, the proposed method also mixs up the query samples by following TransMix and use the synthetic samples to train the model. We divide our contribution into two parts: the transformer based feature extractor and the TransMix based sample mixing. To verify the contribution of each part, the ablation experiments were also conducted. By comparing the proposed method with the method with 3DCNN feature extractor which is used in the work of \cite{33}, we validate the contribution of transformer based feature extractor. At the same time, by comparing the proposed method with that of using CutMix \cite{37} instead of TransMix \cite{36} to mix up query samples, the contribution of TransMix based query sample mixing is also validated.  

Table \ref{Ablation-result} shows the results of the ablation experiments, where Trans denotes the transformer. From the first and second column, it can be seen that when adopting the same mixing technique of CutMix,  better performance are achieved in most cases by adopting transformer as feature extractor. This indicates that compared with the 3DCNN feature extractor, transformer model can learn more point-to-point relationships in the patches. Furthermore, when using transformer as the feature extractor but adopting TransMix instead of CutMix for query sample mixing, there are also better performance which can be seen from the second and third column of Table \ref{Ablation-result}. This proves that compared with CutMix which mixs up the labels of two samples just according to the proportion of synthetic areas, TransMix can generate better mixing labels by using the attention on each pixel which reflect their relative importance. 

\begin{table}[ht]
	\caption{The results of ablation experiment}
	\label{Ablation-result}
	\centering
	\setlength{\tabcolsep}{2mm}{
		\begin{tabular}{|c|c|c|c|}
			\hline
			\rule{0pt}{9pt}
			Var  & CNN+CutMix & Trans+CutMix & Trans+TransMix  \\
			\hline
			\rule{0pt}{9pt}
			Dataset  &	\multicolumn{3}{c|}{IP} \\
			\hline
			\rule{0pt}{9pt}
			OA  & 75.27 +- 3.23 & 76.04 +- 2.96 & \textbf{76.06 +- 2.50}\\
			AA  & 85.32 +- 1.67 & 85.26 +- 1.09 & \textbf{85.34 +- 1.00}\\
			Kappa & 72.23 +- 3.53 & 73.08 +- 3.19 & \textbf{73.10 +- 2.70}\\
			\hline
			\rule{0pt}{9pt}
			Dataset  &	\multicolumn{3}{c|}{PU} \\
			\hline
			\rule{0pt}{9pt}
			OA & 83.63 +- 5.17 & 88.60 +- 3.52 & \textbf{88.83 +- 4.38} \\
			AA & 87.83 +- 2.50 & 90.17 +- 1.67 & \textbf{90.74 +- 2.03} \\
			Kappa & 79.12 +- 5.83 & 85.22 +- 4.30 & \textbf{85.55 +- 5.30} \\
			\hline
			\rule{0pt}{9pt}
			Dataset  &	\multicolumn{3}{c|}{SA} \\
			\hline
			\rule{0pt}{9pt}
			OA & 90.89 +- 2.59 &91.30 +- 2.00& \textbf{91.65 +- 1.59}\\
			AA & 95.10 +- 1.17 &94.99 +- 1.11 & \textbf{95.23 +- 1.00}\\
			Kappa  & 89.88 +- 2.85 & 90.34 +- 2.21& \textbf{90.72 +- 1.75}\\
			\hline
		\end{tabular}
	}
\end{table}

\subsection{Sensitive Analysis of Parameters}
There are several parameters which affect the performance of the proposed method. We have done the experiments to study how sensitive the proposed method is to these parameters. In this section, we report the analysis results of two kinds of parameters, namely the number of transformer encoders and the size of the HSI patch. 

\subsubsection{Effect of the Number of Transformer Encoders}

\begin{figure}[htbp]
	\centering
	\includegraphics[width=9cm,height=6cm]{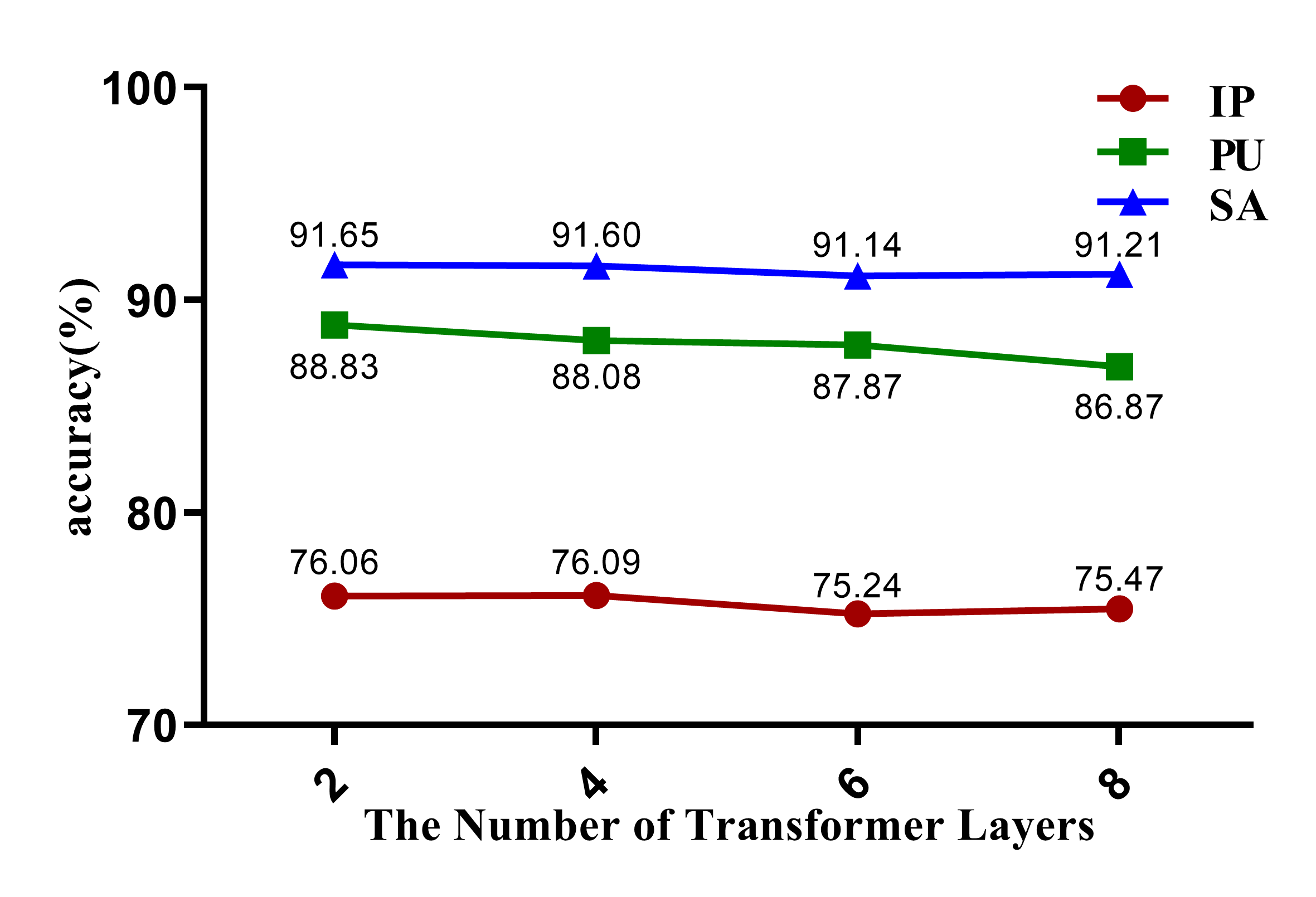}
	\caption{The OA performance changes in different trasformer layers on IP, SA, PU datasets(5 labeled samples)}
	\label{Layers}
\end{figure}

To analyze the impact of the number of transformer encoders on classification performance, we set the number of transformer encoders in the $[2,4,6,8]$ for testing. The changes of the performance of the proposed method on three datasets are shown in Fig. \ref{Layers}. It can be seen that as the number of transformer encoders increases, the performance in OA, AA, and kappa do not significantly improve, and even slightly decreases. This indicates that larger number of transformer encoders  does't mean better performance. This may be because that limited by the number of training samples,  the model has fully learned the classification knowledge in the samples when adopting fewer transformer encoders. It can be seen that when the number of transformer encoders in the model is 2, it achieves the best performance on PU and SA datasets, and achieves the second-best performance on IP dataset. On balance, in order to reduce the scale of model parameters, we finally set the number of transformer encoders to 2 in our experiments.

\subsubsection{Effect of the Patch Size}
\begin{figure}[htbp]
	\centering
	\includegraphics[width=9cm,height=6cm]{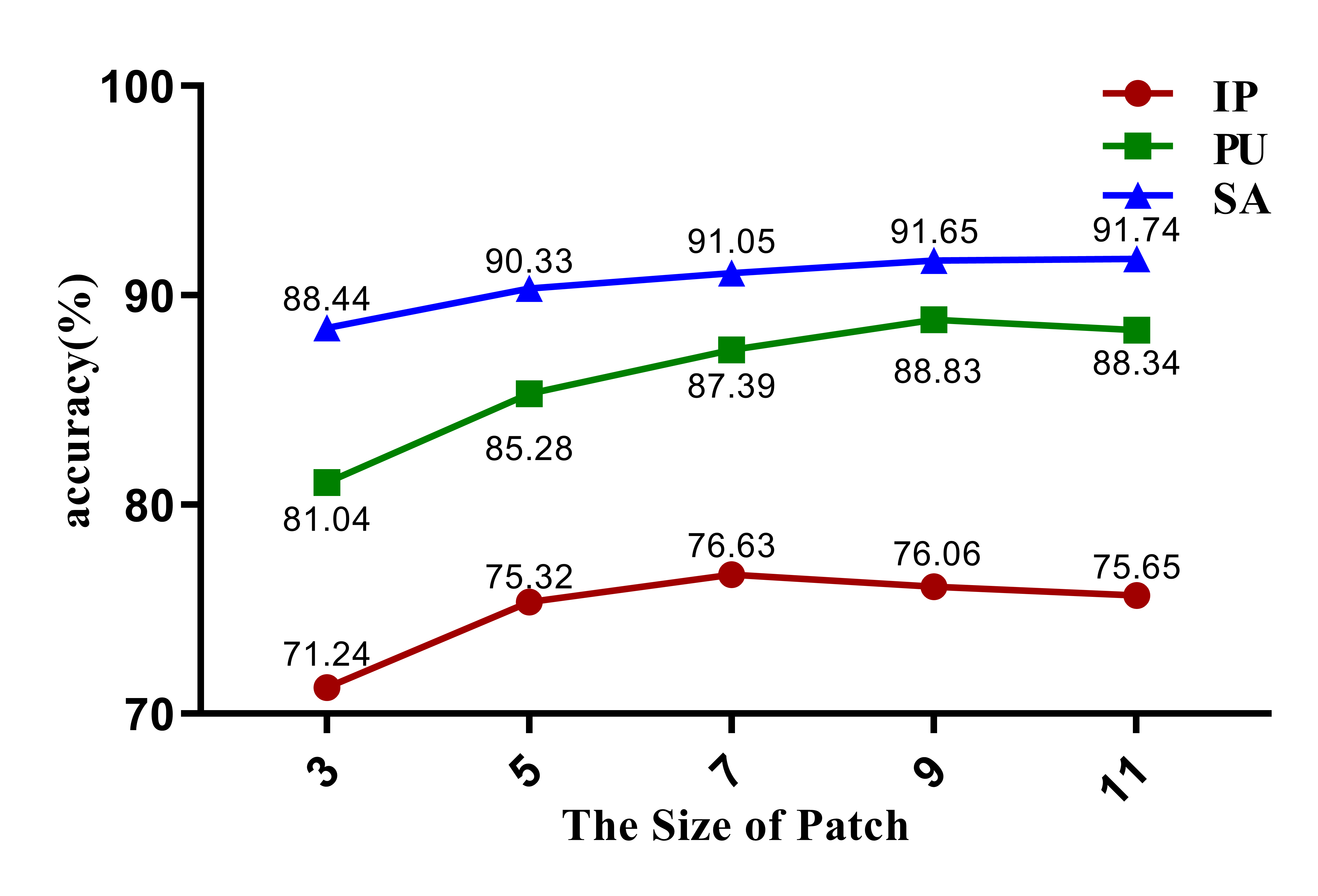}
	\caption{The OA performance changes in different sizes of patch OA on IP, SA, PU datasets(5 labeled samples)}
	\label{Size}
\end{figure}

The proposed method uses transformer to learn the relationship among pixels in the patches and extract the  spatial-spectral joint features from the patches. Then, increasing the patch size will introduce more spatial information. In order to illustrate the influence of patch size on classification performance, we tested the values in $[3, 5, 7, 9, 11]$. The changes of the performance of the proposed method on three datasets are shown in Fig. \ref{Size}. It can be seen that with the change of patch size, the best effect is achieved when the patch size is 7 on IP dataset, while it is better to set the patch size to 11 on PU and SA dataset.  When the size is 9, the proposed method achieves the second-best effect on all three datasets. In order to facilitate the comparison with other models, we set the patch size to 9 in our experiments.

\subsection{Analysis of Parameter and Computational Time}
To further illustrate the performance of the proposed method on time and space consumption, Table \ref{Cost-result} lists the size of model parameters (M) and the inference time including the training and testing time (s) of these comparsion methods. It can be seen that although our APNT method is not the lightest compared to other few-shot methods, it has a faster training speed and competitive testing time. This may be attributed to the parallel execution characteristics possessed by the Transformer model.


\begin{table*}[ht]
	\caption{NUMBER OF PARAMETERS AND COMPUTATIONAL\\TIME OF DIFFERENT METHODS}
	\label{Cost-result}
	\centering
	\setlength{\tabcolsep}{2mm}{
		\begin{tabular}{|c|c|c|c|c|c|c|c|}
			\hline
			\rule{0pt}{9pt}
			Method & CMFSL & Gia-CFSL & MRLF & RPCL & HFSL & APNT* & APNT  \\
			\hline
			\rule{0pt}{9pt}
			Dataset  &	\multicolumn{7}{c|}{IP} \\
			\hline
			\rule{0pt}{9pt}
			Train(s) & 1444.91 & 3697.22 & 2652.73 & 587.21 & 5148.49 & \textbf{213.56} & 459.97\\
			Test(s) & 1.35 & 1.18 & 1.43 & 1.79 & 4.69 & \textbf{0.91} & 1.30\\
			Params(M) & 2.49 & 3.49 & 6.07 & \textbf{0.34} & 12.95 & 3.36 & 3.36\\
			\hline
			\rule{0pt}{9pt}
			Dataset  &	\multicolumn{7}{c|}{PU} \\
			\hline
			\rule{0pt}{9pt}
			Train(s) & 826.91 & 2232.29 & 1843.12 & 278.42 & 2291.59 & \textbf{122.12} & 300.55 \\
			Test(s) & 4.49 & 4.47 & 5.50 & 4.22 & 11.94 & \textbf{2.97} & 4.61 \\
			Params(M) & 2.49 & 3.44 & 6.04 & \textbf{0.30} & 12.37 & 3.32 & 3.32 \\
			\hline
			\rule{0pt}{9pt}
			Dataset  &	\multicolumn{7}{c|}{SA} \\
			\hline
			\rule{0pt}{9pt}
			Train(s) & 1185.34 & 5701.69 & 2796.91 & 510.10 & 6492.53 & \textbf{256.97} & 509.99\\
			Test(s) & \textbf{4.99} & 9.95& 11.80 & 6.46 & 25.68 & 5.10 & 7.08\\
			Params(M) & 2.49 & 3.49 & 6.08 & \textbf{0.34} & 12.97 & 3.35 & 3.35\\
			\hline
		\end{tabular}
	}
\end{table*}

\section{Conclusion}
In this paper, we augment the prototype network with TransMix for few-shot HSI classification. Focusing on the practical problem of low predicting accuracy at boundary pixels, the proposed method uses the transformer as the feature extractor of prototype network, in order to pay different attention to different pixels in the patches adopted for obtaining spatial-spectral joint features. In the meanwhile, it randomly mixs up two patches to imitate the boundary patches which are mixed with multi-class spectral information, and uses these synthetic patches to train the model. It is expected to enlarge the number of hard training samples and enhance the diversity of training samples. And to mix up the labels of two patches, the attention captured by transformer are used to generate better labels for the synthetic patches. The proposed method requires less training time and achieves better results on these datasets widely used for HSI classification experiments. Particularly, in comparative experiments, we found that it  can also achieve good results without using auxiliary datasets for pre-training. This shows that the proposed method can remove the dependence on the auxiliary datasets, and can be easily appied in practice.

\bibliographystyle{IEEEtran}
\bibliography{reference.bib}

\vfill
\end{document}